\documentclass[]{amsart} 

\usepackage{graphicx}
\usepackage[T1]{fontenc} 
\usepackage{ae, aecompl} 
\usepackage{booktabs}
\usepackage{fix-cm}
\usepackage{natbib}

\usepackage{amssymb}
\usepackage{amsmath}
\usepackage{longtable,verbatim}
\usepackage{booktabs}
\usepackage{color, colortbl}
\usepackage{algorithm}
\usepackage{algorithmic}
\usepackage{subfigure}
\usepackage{url}
\usepackage[hyperindex,breaklinks]{hyperref}
\begin{document}

\title[Scalability of using RBMs for Combinatorial Optimization]{Scalability of using Restricted Boltzmann Machines for Combinatorial Optimization}

\author[]{Malte Probst}
\author[]{Franz Rothlauf}
\author[]{Jörn Grahl}
\address[]{Johannes Gutenberg-Universität Mainz\\
Dept.~of Information Systems and Business Administration\\
Jakob-Welder-Weg 9, 55128 Mainz, Germany}
\email{\{probst|rothlauf|grahl\}@uni-mainz.de}
\urladdr{http://wi.bwl.uni-mainz.de}
\begin{abstract}
Estimation of Distribution Algorithms (EDAs) require flexible probability models that can be efficiently learned and sampled. Restricted Boltzmann Machines (RBMs) are generative neural networks with these desired properties. We integrate an RBM into an EDA and evaluate the performance of this system in solving combinatorial optimization problems with a single objective. We assess how the number of fitness evaluations and the CPU time scale with problem size and with problem complexity. The results are compared to the Bayesian Optimization Algorithm, a state-of-the-art EDA. Although RBM-EDA requires larger population sizes and a larger number of fitness evaluations, it outperforms BOA in terms of CPU times, in particular if the problem is large or complex. RBM-EDA requires less time for model building than BOA. These results highlight the potential of using generative neural networks for combinatorial optimization.
\end{abstract}


\maketitle

\section{Introduction}
\label{intro}
Estimation of Distribution Algorithms \citep[EDA,][]{Muehlenbein1996,larranaga2002estimation} are metaheuristics for combinatorial and continuous non-linear optimization. They maintain a population of solutions which they improve over consecutive generations. Unlike other heuristic methods, EDAs do not improve solutions with mutation, recombination, or local search. Instead, they estimate how likely it is that decisions are part of an optimal solution, and try to uncover the dependency structure between the decisions. This information is obtained from the population by the estimation of a probabilistic model. If a probabilistic model generalizes the population well, random samples drawn from the model have a structure and solution quality that is similar to the population itself. Repeated model estimation, sampling, and selection steps can solve difficult optimization problems in theory \citep{Muehlenbein:99a} and in practice \citep{lozano2006towards}. It is important to empirically assess the efficiency of using probability models in EDAs. Simple models, such as factorizations of univariate frequencies, can be quickly estimated from a population, but they cannot represent interactions between the decision variables. As a consequence, EDAs using univariate frequencies cannot efficiently solve complex problems. Using flexible probability models such as Bayesian networks allows complex problems to be solved, but fitting the model to a population and sampling new solutions can be very time-consuming. 

A central goal of EDA research is the identification of probabilistic models that are flexible and can quickly be estimated and sampled. This is also a central topic in the field of machine learning. A recent focus in machine learning is the development of feasible unsupervised learning algorithms for generative neural networks. These algorithms and models can learn complex patterns from high-dimensional datasets. Moreover, generative neural networks can sample \textit{new data} based on the associations that they have learned so far, and they can be fit to data (e.g., to a population of solutions) in an unsupervised manner. This makes them potentially useful for EDAs. Some of these models can also be ``stacked'' on several layers and be used as building blocks for ``deep learning''. 

In this paper, we focus on Restricted Boltzmann Machines \citep[RBM,][]{Smolensky1986,Hinton2002}. RBMs are a basic, yet powerful, type of generative neural network where the connections between the neurons form a bipartite graph (Section \ref{rbm} explains RBMs in detail). Due to a recent breakthrough \citep{Hinton2006}, training RBMs is computationally tractable. They show impressive performance in classic machine learning tasks such as image or voice recognition \citep{dahl2012context}. 

Given these successes, it is not surprising that researchers have integrated RBMs and similar models into EDAs and studied how these systems perform in optimization tasks. \cite{zhang2000bayesian} used a Helmholtz Machine in an EDA. Helmholtz Machines are predecessors of RBMS. Due to their limited performance, they are nowadays widely discarded. \cite{zhang2000bayesian} evaluated their EDA by comparing it to a simple Genetic Algorithm \citep{goldberg1989genetic}. They did not study the scaling behavior for problems of different sizes and complexity. In a series of recent papers, \citet{Tang2010RBM-EDA, Shim2010, shin-tan2012} and \citet{Shim2013} studied EDAs that use RBMs. These works are similar to ours in that an RBM is used inside an EDA. An important difference is that they considered problems with multiple objectives. Also, they hybridized the EDA with particle swarm optimization. Thus, it is unknown if using and RBM in an EDA leads to competitive performance in single-objective combinatorial optimization. 

Therefore, in this paper, we raise the following questions:
\begin{enumerate}
\item How efficient are EDAs that use RBMs for single-objective combinatorial optimization?
\item How does the runtime scale with problem size and problem difficulty?
\item Is the performance competitive with the state-of-the-art?
\end{enumerate}

To answer these questions, we integrated an RBM into an EDA (the RBM-EDA) and conducted a thorough experimental scalability study. We systematically varied the difficulty of two tunably complex single-objective test  problems (concatenated trap functions and NK landscapes), and we computed how the runtime and the number of fitness evaluations scaled with problem size. We then compared the results those obtained for the Bayesian Optimization Algorithm \citep[BOA,][]{Pelikan1999,Pelikan2005}. BOA is a state-of-the-art EDA. 

RBM-EDA solved the test problems in polynomial time depending on the problem size. Indeed, concatenated trap functions can only be solved in polynomial time by decomposing the overall problem into smaller parts and then solving the parts independently. RBM-EDA's polynomial scalability suggests that RBM-EDA recognized the structure of the problem correctly and that it solved the sub-problems independently from one another. The hidden neurons of the RBM (its latent variables) captured the building blocks of the optimization problem. The runtime of RBM-EDA scaled better than that of BOA on trap functions of high order and NK landscapes with large $k$. RBM-EDA hence appears to be useful for complex problems. It was mostly faster than BOA if instances were large.

The paper is structured as follows: In Section \ref{eda}, we introduce Estimation of Distribution Algorithms and the Bayesian Optimization Algorithm. In section \ref{rbm}, we introduce Restricted Boltzmann Machines, show how an RBM samples new data, and describe how an RBM is fit to given data. Section \ref{rbm-eda} describes RBM-EDA. The test functions, the experimental design and the results are presented in Section \ref{experiments}. Section \ref{conclusion} concludes the paper.

\section{Estimation of Distribution Algorithms}
\label{eda}

We introduce Estimation of Distribution Algorithms (Section \ref{eda-basic}) and the Bayesian Optimization Algorithm (Section \ref{boa}).

\subsection{Estimation of Distribution Algorithms}
\label{eda-basic}

EDAs are population-based metaheuristics \citep{Muehlenbein1996,Muehlenbein:99a,Pelikan:99e,Larranaga:99x}. Similar to Genetic algorithms \citep[GA,][]{Holland:75a,goldberg1989genetic}, they evolve a population of solutions over a number of generations by means of selection and variation.

\begin{algorithm}[bp]
\caption{Estimation of Distribution Algorithm}
\label{alg-eda}
\begin{algorithmic}[1]
\STATE \textbf{Initialize} Population $P$
\WHILE {not converged}
\STATE    $P_{parents}$ $\leftarrow$ \textbf{Select} high-quality solutions from $P$ based on their fitness
\STATE    $M$ $\leftarrow$ \textbf {Build} a model  estimating the  (joint) probability distribution of $P_{parents}$ 
\STATE    $P_{candidates}$ $\leftarrow$ \textbf{Sample} new candidate solutions from $M$
\STATE    $P$ $\leftarrow$ $P_ {parents}\cup P_ {candidates}$
\ENDWHILE
\end{algorithmic}
\end{algorithm}

Algorithm \ref{alg-eda} outlines the basic functionality of an EDA. After initializing a population $P$ of solutions, the EDA runs for multiple generations. In each generation, a selection operator selects a subset $P_{parents}$ of high-quality solutions from $P$. $P_{parents}$ is then used as input for the variation step. In contrast to a GA, which creates new individuals using recombination and mutation, an EDA builds a probabilistic model $M$ from $P_{parents}$, often by estimating their (joint) probability distribution. Then, the EDA draws samples from $M$ to obtain new candidate solutions. Together with $P_{parents}$, these candidate solutions constitute $P$ for the next generation. The algorithm stops after the population has converged or another termination criterion is met.

EDA variants mainly differ in their probabilistic models $M$. The models describe dependency structures between the decisions variables with different types of probability distributions. Consider a binary solution space with $n$ decision variables. A naive EDA could attempt to store a probability for each solution. $M$ would contain $2 ^n$ probabilities. This could be required if all variables depended on each other. However, storing $2^n$ probabilities is computationally intractable for large $n$. If some decision variables are, however, independent from other variables, then the joint distribution could be factorized into products of marginal distributions and the space required for storing $M$ shrinks. If all variables are independent, only $n$ probabilities have to be stored. In most problems, some variables are independent of other variables but the structure of the dependencies is unknown to those who want to solve the problem. Hence, model building consists of finding a network structure that matches the problem structure and estimating the model's parameters.

Simple models like the Breeder Genetic Algorithm \citep{Muehlenbein1996} or population-based incremental learning \citep{baluja1994population} use univariate, fully factorized probability distributions with a vector of activation probabilities for the variables and choose to ignore dependencies between decision variables. Slightly more complex approaches like the bivariate marginal distribution algorithm use bivariate probability distributions which model pairwise dependencies between variables as trees or forests \citep{pelikan1999bivariate}. More complex dependencies between variables  can be captured by models with multivariate interactions, like the Bayesian Optimization Algorithm \citep[][see section \ref{boa}]{Pelikan1999} or the extended compact GA \citep{Harik:99}. Such multivariate models are better suited for complex optimization problems. Univariate models can lead to an exponential growth of the required number of fitness evaluations \citep{Pelikan1999, Pelikan2005}. However, the computational effort to build a model $M$ increases with its complexity and representational power. Many algorithms use probabilistic graphical models with directed edges, i.e., ~Bayesian networks, or undirected edges, i.e., ~Markov random fields \citep{larranaga2012review}. 

\subsection{Bayesian Optimization Algorithm}
\label{boa}

The Bayesian Optimization Algorithm is the state-of-the-art EDA optimization algorithm for discrete optimization problems. It was been proposed by \cite{Pelikan1999} and has been heavily used and researched since then \citep[]{pelikan2003hierarchical,  Pelikan2008techreport, abdollahzadeh2012bayesian}. 

BOA uses a Bayesian network for modeling dependencies between variables. Decision variables correspond to nodes and dependencies between variables correspond to directed edges. As the number of possible network topologies grows exponentially with the number of nodes, BOA uses a greedy construction heuristic to find a network structure $G$ to model the training data. Starting from an unconnected (empty) network, BOA evaluates all possible additional edges, adds the one that maximally increases the fit between the model and selected individuals, and repeats this process until no more edges can be added. The fit between selected individuals and the model is measured by the Bayesian Information Criterion \citep[BIC,][]{schwarz1978estimating}. BIC is based on the conditional entropy of nodes given their parent nodes and correction terms penalizing complex models. BIC assigns each network $G$ a scalar score
\begin{equation}
\label{eq-bic}
BIC(G)=\sum_{i=1}^{n}\left(-H\left(X_i|\Pi_i\right)N-2^{|\Pi_i|}\frac{\log_2(N)}{2}\right),
\end{equation}
where $n$ is the number of decision variables, $N$ is the sample size (i.e., the number of selected individuals), $\Pi_i$ are the predecessors of node $i$ in the Bayesian network ($i$'s parents), and $|\Pi_i|$ is the number of parents of node $i$. The term $H(X_i|\Pi_i)$ is the conditional entropy of the $i$'th decision variable $X_i$ given its parental nodes, $\Pi_i$, defined as
\begin{equation}
H(X_i|\Pi_i)=-\sum_{x_i,\pi_i}{p(x_i,\pi_i) \log_2 p(x_i|\pi_i)},
\label{eq:h}
\end{equation}
\noindent where $p(x_i, \pi_i)$ is the observed probability of instances where $X_i=x_i$ and $\Pi_i=\pi_i$, and $p(x_i|\pi_i)$ is the conditional probability of instances where $X_i=x_i$ given that $\Pi_i=\pi_i$. The sum in (\ref{eq:h}) runs over all possible configurations of $X_i$ and $\Pi_i$.
The BIC score depends only on the conditional entropy of a node and its parents. Therefore, it can be calculated independently for all nodes. If an edge is added to the Bayesian network, the change of the BIC can be computed quickly. The term $-2^{|\pi_i|}\ldots$ in (\ref{eq-bic}) penalizes model complexity.
BOAs greedy network construction algorithm adds the edge with the largest gain in $\mbox{BIC}(G)$ until no more edges can be added. Edge additions resulting in cycles are not considered. 

After the network structure has been learned, BOA calculates the conditional activation probability tables for each node. Once the model structure and conditional activation probabilities are available, BOA can produce new candidate solutions by drawing random values for all nodes in topological order.

\section{Restricted Boltzmann Machines and the RBM-EDA}
\label{rbm}
Restricted Boltzmann Machines \citep{Smolensky1986} are stochastic neural networks that are successful in areas such as image classification, natural language processing, or collaborative filtering \citep{Dahl2010,Hinton2006,Salakhutdinov07}. In this section, we describe the structure of Restricted Boltzmann Machines (Section \ref{rbm-structure}), show how an RBMs can sample new data (Section \ref{rbm-sampling}), and how contrastive divergence learning is used to model the probability distribution of given data (Section \ref{rbm-training}). Finally, we describe RBM-EDA, an EDA that uses an RBM as its probabilistic model (Section \ref{rbm-eda}). 
\subsection{Structure of RBMs}
\label{rbm-structure}
\begin{figure}
\begin{center}
\centerline{\includegraphics[width=0.4\columnwidth]{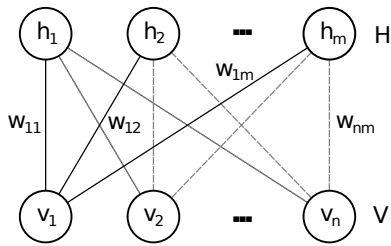}}
\caption{A Restricted Boltzmann Machine as a graph. The visible neurons $v_i$ ($i\in{1..n}$) can hold a data vector of length $n$ from the training data. In the EDA context, $V$ represents decision variables. The hidden neurons $h_j$ ($j \in{1..m}$) represent $m$ features. Weight $w_{ij}$ connects $v_i$ to $h_j$.}
\label{fig-rbm}
\end{center}
\end{figure}

Figure \ref{fig-rbm} illustrates the structure of an RBM. We denote $V$ as the input (or ``visible'') layer.  $V$ holds the input data represented by $n$ binary variables $v_i, i= 1,2,\ldots,n$.  The $m$ binary neurons $h_j, j=1,2,\ldots,m$ of the hidden layer $H$ are called feature detectors as they are able to model patterns in the data. A weight matrix $W$ holds weights $w_{i,j}\in \mathbb{R}$ between all neurons $v_i$ and $h_j$. Together, $V$, $H$, and $W$ form a bipartite graph. The weights $W$ are undirected. An RBM forms a Markov random field. 

In the sampling and training phase, each neuron in $V$ and $H$ makes stochastic decisions about whether it is active (its value then becomes 1) or not (its value then becomes 0). Therefore, it collects inputs from all neurons to which it is directly connected. $W$ determines the strengths of the inputs.

An RBM encodes a joint probability distribution $P(V,H)$. In the sampling phase, a configuration of $V$ and $H$ is thus sampled with probability $P(V,H)$ \citep{Smolensky1986}. In the training phase, the weights $W$ are adapted such that the marginal probability $P(V)$ approximates the probability distribution of the training data. Training and sampling are tractable because of the bipartite structure of the RBM. Hence, it is not necessary to know the problem structure beforehand.

\subsection{Sampling}
\label{rbm-sampling}
The goal of the sampling phase is to generate new values for the neurons in the visible layer $V$ according to $P(V,H)$. This is straightforward if the activations of the neurons in the hidden layer $H$ are known. In this case, all $v_i$ are independent of each other and the conditional probability that $v_i$ is active is simple to compute. The conditional probability $P(v_i=1|H)$ that the visible neuron $v_i$ is active, given the hidden layer $H$ is calculated as
\begin{equation}
\label{eq-Pv}
P(v_i=1|H)=\text{sigm}\left(\sum_{j}w_{ij}h_j\right),
\end{equation}
where  $\text{sigm}(x)=\frac{1}{1+e^{-x}}$ is the logistic function.\footnote{In addition, all neurons are connected to a special ``bias'' neuron, which is always active and works like an offset to the neuron's input. Bias weights are treated like normal weights during learning. Due to brevity, we omitted the bias terms throughout the paper. For details, see \cite{Hinton2006}.} Analogously, given the activations of the visible layer $V$, the conditional probability $P(h_j=1|V)$ for the hidden neurons $H$ is calculated as 
\begin{equation}
\label{eq-Ph}
P(h_j=1|V)=\text{sigm}\left(\sum_{i}w_{ij}v_i\right).
\end{equation}
Although the two conditional distributions $P(V|H)$ and $P(H|V)$ are simple to compute, sampling from the joint probability distribution $P(V,H)$ is much more difficult, as it usually requires integrating over one of the conditional distributions. An alternative is to use Gibbs sampling, which approximates the joint distribution $P(V,H)$ from the conditional distributions. Gibbs sampling starts by assigning random values to the visible neurons. Then, it iteratively samples from $P(H|V)$ and $P(V|H)$, respectively, while assigning the result of the previous sampling step to the non-sampled variable. Sampling in the order of $V\rightarrow H\rightarrow V\rightarrow H\rightarrow V \rightarrow...$ forms a Markov chain. Its stationary distribution is identical to the joint probability distribution $P(V,H)$ \citep{GemanGeman1993}. The quality of the approximation increases with the number of sampling steps. If Gibbs sampling is started with a $V$ that has a high probability under the stationary distribution, only a few sampling steps are necessary to obtain a good approximation.

\subsection{Training}
\label{rbm-training}
In the training phase, the RBM adapts the weights $W$ such that $P(V)$ approximates the distribution of the training data. An effective approach for adjusting the weights of an RBM is contrastive divergence (CD) learning \citep{Hinton2002}.
CD learning maximizes the log-likelihood of the training data under the model, $\log(P(V))$, by performing a stochastic gradient ascent. The main element of CD learning is  Gibbs sampling.

For each point $V$ in the training data, CD learning updates $w_{ij}$ in the direction of $-\frac{\partial \log(P(V))}{\partial w_{ij}}$.  This partial derivative is the difference of two terms usually referred to as positive and negative gradient,  $\Delta^{\text{pos}}_{ij}$ and $\Delta^{\text{neg}}_{ij}$ \citep{Hinton2002}. The total gradient $\Delta w_{ij}$ is
\begin{equation}
\begin{aligned}
\label{eq-cd}
\Delta w_{ij}&=\quad\Delta^{\text{pos}}_{ij}    &-&\quad\Delta^{\text{neg}}_{ij} \\
              &=\;<v_i\cdot h_j>^{\text{data}} &- &\;<v_i\cdot h_j>^{\text{model}},
\end{aligned}
\end{equation}
where $<x>$ is the expected value of $x$.
$\Delta^{\text{pos}}_{ij}$ is the expected value of $v_ih_j$ when setting the visible layer $V$ to a data vector from the training set and sampling $H$  according to (\ref{eq-Ph}). $\Delta^{\text{pos}}_{ij}$ increases the marginal probability $P(V)$ of the data point $V$. In contrast, $\Delta^{\text{neg}}_{ij}$ is the expected value of a configuration sampled from  the joint probability distribution $P(V,H)$, which is approximated by Gibbs sampling. If $P(V)$ is equal to the distribution of the training data, in the expectation, the positive and negative gradient equal each other and the total gradient becomes zero.

Calculating $\Delta^{\text{neg}}_{ij}$ exactly is infeasible since a high number of Gibbs sampling steps is required until the RBM is sampling from its stationary distribution. Therefore, CD estimates $\Delta^{\text{neg}}_{ij}$ by using two approximations. First, CD initializes the Markov chain with a data vector from the training set, rather than using an unbiased, random starting point. Second, only a small number of sampling steps is used. We denote CD using $N$ sampling steps as CD-$N$. CD-$N$ approximates the negative gradient $\Delta^{\text{neg}}_{ij}$ by initializing the sampling chain to the same data point $V$ which is used for the calculation of $\Delta^{\text{pos}}_{ij}$. Subsequently, it performs $N$ Gibbs sampling steps. Note that the first half-step $V\rightarrow H$ has, in practice, already been performed during the calculation of  $\Delta^{\text{pos}}_{ij}$. Despite using these two approximations, CD-$N$ usually works well \citep{Hinton2006}.

Algorithm \ref{alg-cd} describes contrastive divergence for $N=1$  (CD-1). For each training vector, $V$ is initialized with the training vector and $H$ is sampled according to (\ref{eq-Ph}). This allows the calculation of $\Delta^{\text{pos}}_{ij}$ as $v_ih_j$. Following this, two additional sampling steps are carried out: First, we calculate the ``reconstruction'' $\hat{V}$ of the training vector as in (\ref{eq-Pv}). Subsequently, we calculate the corresponding hidden probabilities $P(\hat{h_j}=1|\hat{V})$. Now, we can approximate $\Delta^{\text{neg}}_{ij}$  as $\hat{v}_i \cdot P(\hat{h_j}|\hat{V})$ and obtain $\Delta w_{ij}$. Finally, we update the weights as
\begin{equation}
\label{eq-update}
w_{ij}:=w_{ij}+\alpha \cdot \Delta w_{ij}
\end{equation}
where $\alpha\in(0,\ldots, 1)$ is a learning rate defined by the user. This procedure repeats for several epochs, i.e., passes through the training set. Usually, CD is implemented in a mini-batch fashion. That is, we calculate $\Delta w_{ij}$ in  (\ref{eq-update}) for multiple training examples at the same time, and subsequently use the average gradient to update $w_{ij}$. This reduces sampling noise and makes the gradient more stable \citep{bishop2006pattern, Hinton2006}

\begin{algorithm}[bp]
\begin{algorithmic}[1]
\caption{Pseudo code for a training epoch using CD-1}
\label{alg-cd}
\FOR {all training examples}
\STATE  $V$ \quad\, $\leftarrow$            set $V$ to the current training example
\STATE  $H$ \quad\, $\leftarrow$            sample $H|V$, i.e.~set $h_j$ to 1 with $P(h_j=1|V)$ from (\ref{eq-Ph}) 
\STATE  $\Delta_{ij}^{\text{pos}}=v_ih_j$
\STATE  $\hat{V}$ \quad\, $\leftarrow$            sample "reconstruction" $\hat{V}|H$, using (\ref{eq-Pv})
\STATE  $\hat{H}$ \quad\, $\leftarrow$            calculate $P(\hat{H}|\hat{V})$ as in (\ref{eq-Ph}) 
\STATE  $\Delta_{ij}^{\text{neg}}=\hat{v}_i\cdot P(\hat{h}_j|\hat{V})$
\STATE  $\Delta w_{ij}$ $\leftarrow$ calculate all $\Delta w_{ij}$ as in (\ref{eq-cd})
\STATE  $w_{ij}$ \;\, $\leftarrow$            update all weights according to (\ref{eq-update})
\ENDFOR
\end{algorithmic}
\end{algorithm}

\subsection{Restricted Boltzmann EDA}
\label{rbm-eda}
This section describes how we used an RBM in an EDA. The RBM should model the properties of promising solutions and then be used to sample new candidate solutions. In each generation of the EDA, we trained the RBM to model the probability distribution of the solutions which survived the selection process. Then, we sampled candidate solutions from the RBM and evaluate their fitness. 

We chose to use the following parameters: The number $m$ of hidden neurons was set to be half the number $n$ of visible neurons (i.e.~half the problem size). Standard values were used for the parameters that control the training and sampling phase \citep{Hinton2010techreport}.  

\paragraph{Sampling parameters} When sampling new candidate solutions from the RBM, we used the individuals in $P_{parents}$ to initialize the visible neurons close to the stationary distribution. Subsequently, we performed 25 full Gibbs sampling steps.

\paragraph{Training parameters} The learning rate $\alpha$ was set to 0.05 and 0.5 for the weights and biases, respectively. 
We applied standard momentum \citep{qian1999momentum}, which adds a fraction $\beta$ of the last parameter update to the current update, making the update less prone to fluctuations caused by noise, and dampening oscillations. We started with $\beta=0.5$ and increased it during an EDA run to $\beta=0.8$  (see below). 

We used weight decay (L2 regularization) to improve generalization. Weight decay adds the term $-0.5*\gamma*W^2$ to the RBM's optimization objective. The partial gradient $\Delta w_{ij}$ in Equation (\ref{eq-cd}) thus includes the term $-\gamma*w_{ij}$, which decays large weights and thereby reduces overfitting. The weight cost $\gamma$ determines the strength of the decay.
We chose  $\gamma=0.0001$.

We used contrastive divergence learning with one Gibbs sampling step (CD-$1$) and a mini-batch size of 100. Furthermore, we  initialized the visible biases with $\log\left(\frac{P(v_i=1)}{1-P(v_i=1)}\right)$, where $P(v_i=1)$ is the probability that the visible neuron $v_i$ is set to one in the training set \citep[see][]{Hinton2010techreport}. This initialization speeds up the training process.

\paragraph{Parameter control}


We applied a simple parameter control scheme for the learning rate $\alpha$, the momentum $\beta$, and the number of epochs.
The scheme was based on the reconstruction error $e$. The reconstruction error is the difference between a training vector $V$ and its reconstruction $\hat{V}$ after a single step of Gibbs sampling (see lines 2 and 7 in Algorithm \ref{alg-cd}).\footnote{CD learning does not minimize the reconstruction error $e$ but maximizes $P(V)$, which cannot be calculated exactly tractably. As an alternative, the reconstruction error $e$ is usually a good approximation for how good the model can (re-)produce the training data.} $e$ usually decreases with the number of epochs. Every second epoch $t\in{1,\ldots,T}$, we calculated for a fixed subset $s$ of the training set $S$ the relative difference
$$
e_t^s=1/{|s|}\sum_{j\in s}\sum_i|v_i-\hat{v}_i|/n
$$ 
 between $v$ and $\hat{v}$.
We measured the decrease $\gamma$ of the reconstruction error in the last 25\% of all epochs as $\gamma=\frac{e^S_{0.75t}-e^S_t}{e^S_0-e^S_t}$. $\gamma$ was then used to adjust the learning parameters.
The learning rate $\alpha$ was initialized in the first epoch with $\alpha=0.05$. As soon as $\gamma<.05$, $\alpha$ was decreased to $0.025$. In the first epoch we set the momentum $\beta$ to $0.5$. As soon as $\gamma<0.1$ we increased $\beta$ to $\beta=0.8$. 

\noindent We used two self-adaptive termination criteria for stopping the training phase. We stopped training if $\gamma<0.01$. The rationale behind this was that the RBM has learned the relevant dependencies between the variables, and further training was unlikely to improve the model considerably. Furthermore, we stopped the training if the RBM was overfitting, i.e., learning noise instead of problem structure. Therefore, we split the original training set into a training set $S$ containing 90\% of all samples and a validation set $S'$ containing the remaining 10\%. We trained the RBM only for the solutions in $S$ and, after each epoch, calculated the reconstruction error $e^{S}$ and $e^{S'}$ for the training and validation set $S$ and $S'$, respectively. We stopped the training phase as soon as $\frac{|e^{S}-e^{S'}|}{e^{S'}}\geq 0.02$ (i.e., the absolute difference between the reconstruction errors was larger than 2\%). 

\section{Experiments}
\label{experiments}

We describe the test problems (Section \ref{sec:test-problems}) and our experimental design (Section \ref{experiment}). The results are presented in Section \ref{results}.

\subsection{Test Problems}
\label{sec:test-problems}

We evaluated the performance of RBM-EDA on onemax, concatenated deceptive traps \citep{Ackley:87}, and NK landscapes \citep{kauffman1989nk}. All three are standard benchmark problems. Their difficulty depends on the problem size, i.e., problems with more decision variables are more difficult. Furthermore, the difficulty of concatenated deceptive trap functions and NK landscapes is tunable by a parameter. All three problems are defined on binary strings of fixed length. 
 
The onemax problem assigns a binary solution $x$ of length $l$ a fitness value $f=\sum^l_{i=1} x_i$, i.e., the fitness of $x$ is the number of ones in $x$. The onemax function is rather simple. It is unimodal and can be solved by a deterministic hill climber. 

A trap function is defined on  a binary solution $x$ of length $k$. It assigns a solution $x$ a fitness
$$
f^k(x) = 
\begin{cases} 
k &\mbox{if } \sum_{i}{x_i}=k, \mbox{ and}  \\
k-(\sum_{i}{x_i}+1) & \mbox{otherwise.}
\end{cases}
$$
The optimal solution consists of all ones. Trap functions are difficult to solve because the second-best solution consists of all zeros and the structure of the function is deceptive. It leads search methods away from the global optimum towards the second-best solution.

A \textit{concatenated} trap function places $o$ trap functions of order $k$ on a single bitstring $x$ of length $o\cdot k$. Its fitness is calculated as $f(x)=\sum_{i=1}^o f^k_i$ \citep{Ackley:87}. The problem difficulty increases with $k$  as well as with $o$. Concatenated traps are decomposable. Dependencies exist between the $k$ variables of a trap function, but not between variables in the $o$ different trap functions \citep{deb1991analyzing}. 

NK landscapes are defined by two parameters $N$ and $k$ and $N$ fitness components $f^{N}_i$ \citep{kauffman1989nk}. A solution vector $x$ consists of $l=N$ bits. The bits are assigned to $N$ overlapping subsets, each of size $k+1$. The fitness of a solution is the sum of $N$ fitness components. Each component $f^{N}_i$ depends on the value of the corresponding variable $x_i$ as well as $k$ other variables. Each $f^{N}_i$ maps each possible configurations of its $k+1$ variables to a fitness value. The overall fitness function is therefore
$$
f(x)=1/N\sum_{i=1}^Nf^N_i(x_i,x_{i1},\ldots,x_{iK}).
$$ 
\noindent Each decision variable usually influences several $f_i^N$. These dependencies between subsets make NK landscapes non-decomposable. The problem difficulty increases with $k$. $k=0$ is a special case where all decision variables are independent and the problem reduces to onemax. 

\subsection{Experimental Setup}
\label{experiment}
We adopted an experimental setup similar to \cite{Pelikan2008techreport}. We studied the performance of RBM-EDA on the onemax function, on concatenated deceptive traps with traps size $k\in\{4,5,6\}$, and NK landscapes with $k\in \{2,3,4,5\}$. We report results for BOA so that RBM-EDA can be compared to the state-of-the-art. 

Both EDAs used tournament selection without replacement of size two \citep{Miller95geneticalgorithms}. 
We used bisection to determine the smallest population size for which a method solved a problem to optimality. For onemax and deceptive traps, we required each method to find the optimal solution in 30 out of 30 independent runs. For NK landscapes, we used 25 randomly chosen problem instances per size and determined the population size that solved five out of five independent runs of each instance. 

We report the average number of fitness evaluations that were necessary to solve the problem to optimality. In addition, we report average CPU running times. CPU running time includes the time required for fitness calculation, the time required for model building (either building the Bayesian network or training the RBM), the time required for sampling new solutions, and the time required for selection. We also report CPU times even though previous EDA research mostly ignored the time required for model building and sampling. 

We implemented RBM-EDA and BOA in Java. The experiments were conducted on a single core of an AMD Opteron 6272 processor with 2,100 MHz. The JBLAS library was used for the linear algebra operations of the RBM.
\subsection{Results}
\label{results}
\begin{figure}
  \begin{center}
    \subfigure
    {       \includegraphics[width=0.45\linewidth]{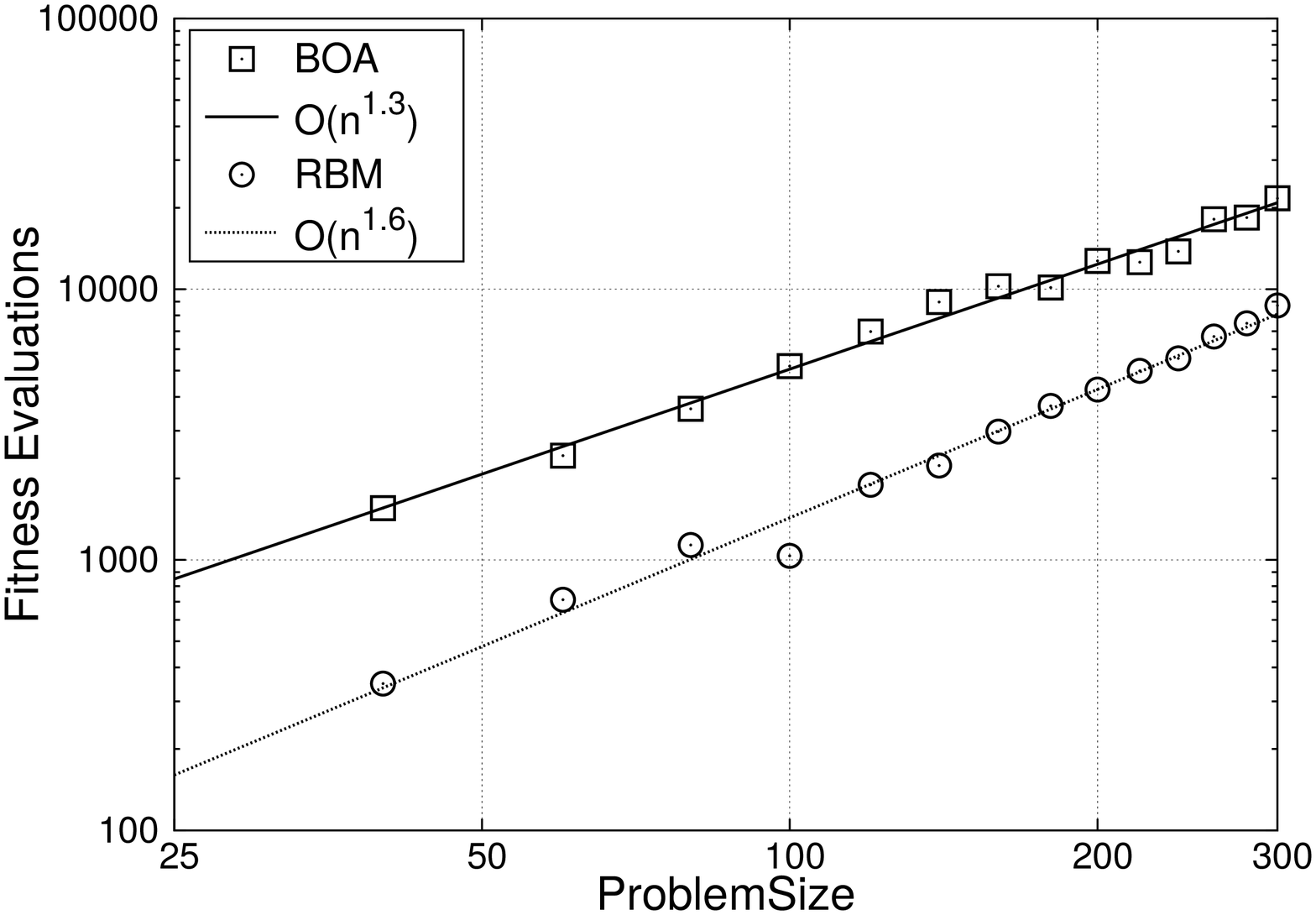}\label{fig-onemax-eval}
    }
    \subfigure
    {
      \includegraphics[width=0.45\linewidth]{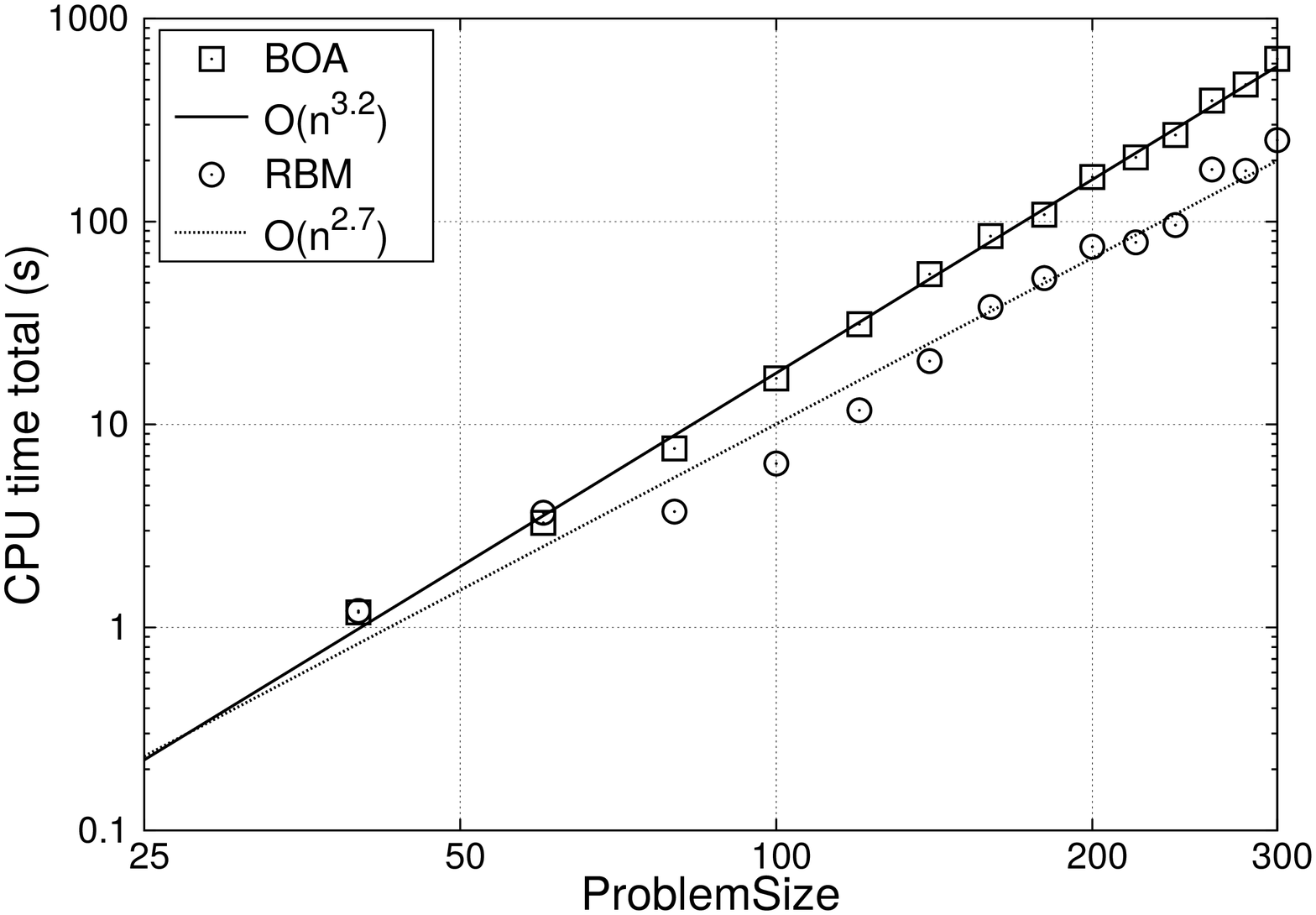}
      \label{fig-onemax-cpu}
    }
    \caption{Number of evaluations (left) and CPU time (right) over problem size $l$ for the onemax problem.\label{fig-onemax}}
  \end{center}
\end{figure}

\begin{figure}
  \begin{center}
    \subfigure[$k=4$\label{fig-4-traps}]
    {
      \includegraphics[width=0.45\linewidth]{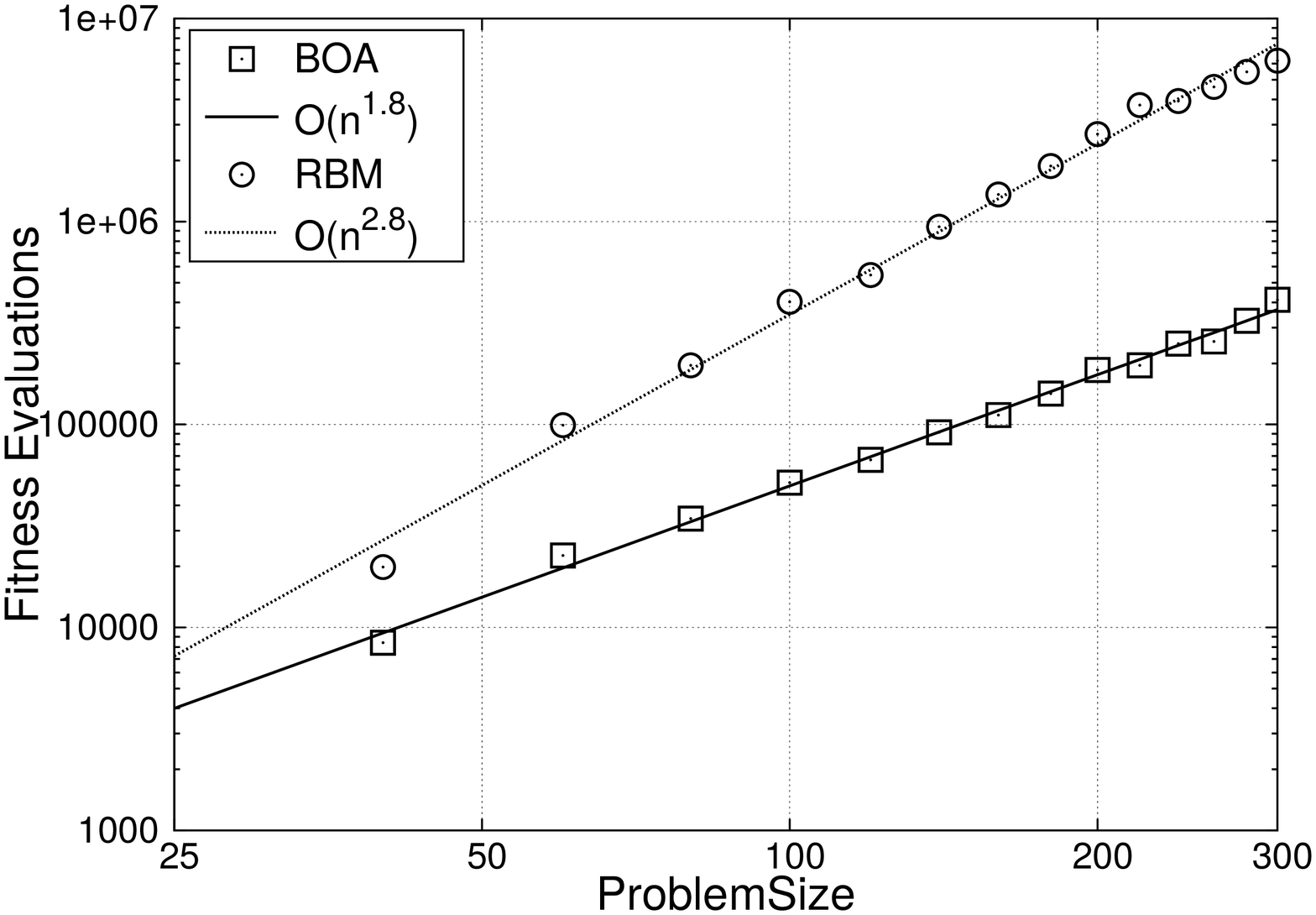}\label{fig-4-trap-eval}
      \includegraphics[width=0.45\linewidth]{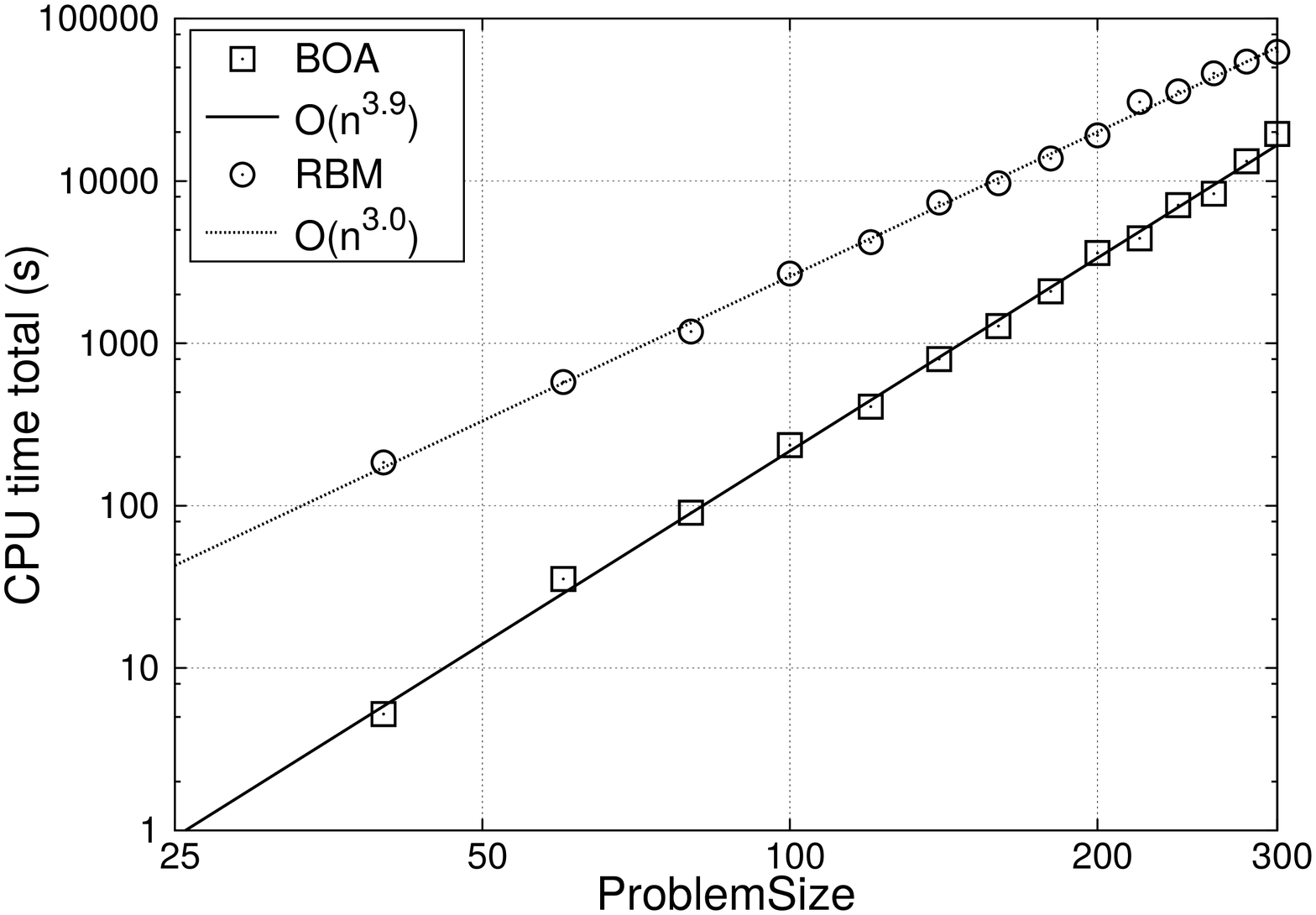} \label{fig-4-trap-cpu}
    }
    \subfigure[$k=5$\label{fig-5-traps}]
    {
      \includegraphics[width=0.45\linewidth]{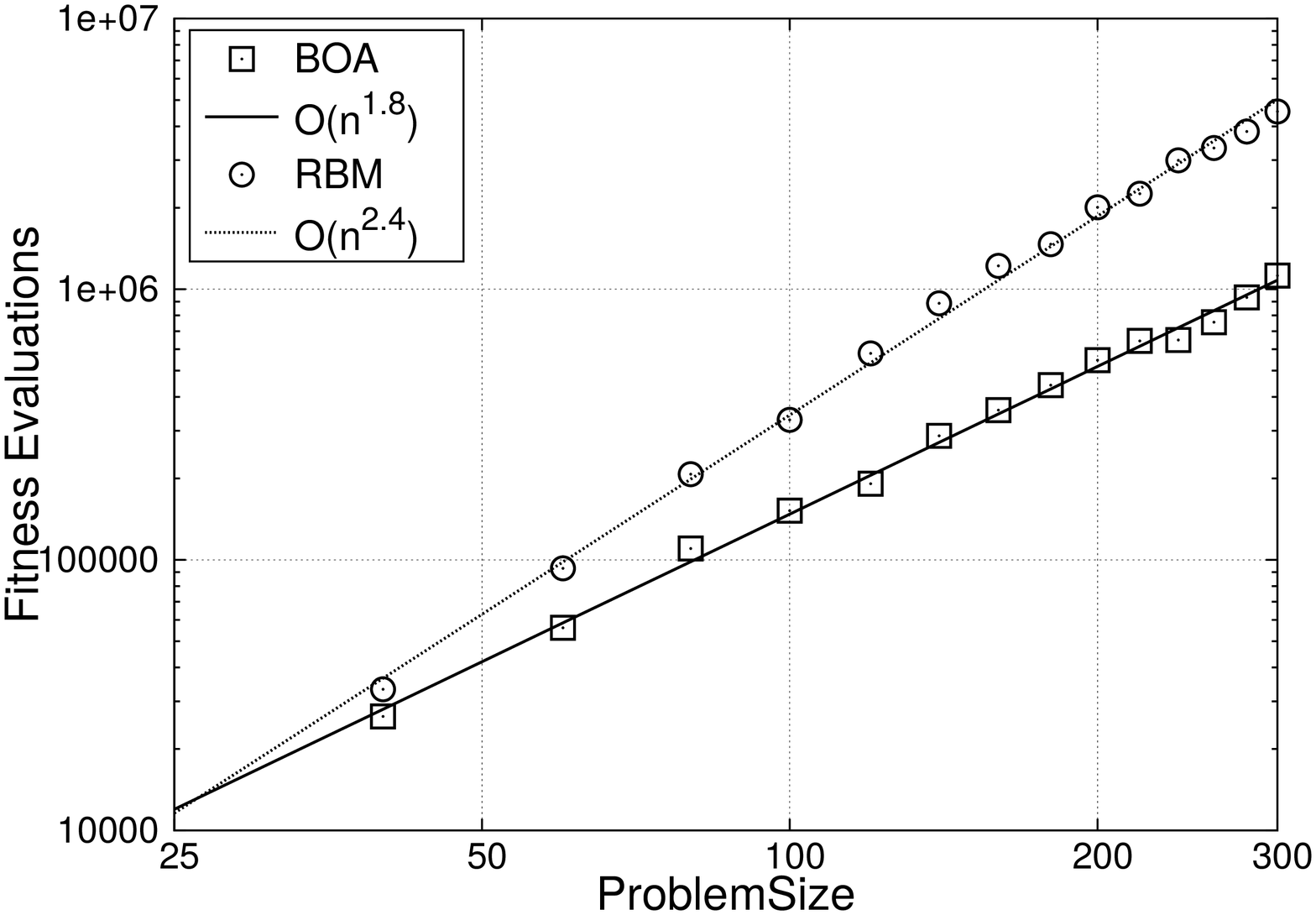}\label{fig-5-trap-eval}
      \includegraphics[width=0.45\linewidth]{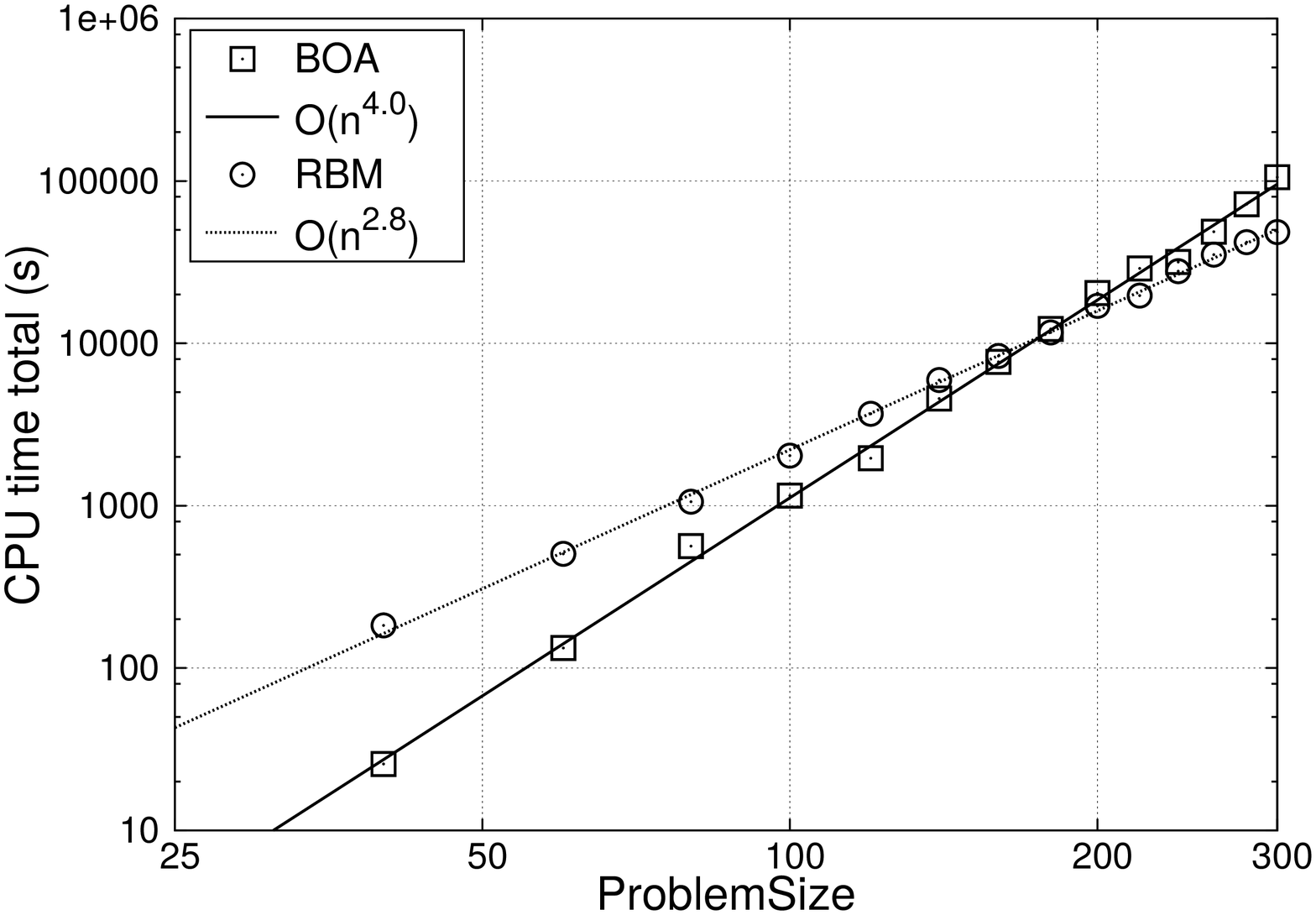} \label{fig-5-trap-cpu}
    }
    \subfigure[$k=6$\label{fig-6-traps}]
    {
      \includegraphics[width=0.45\linewidth]{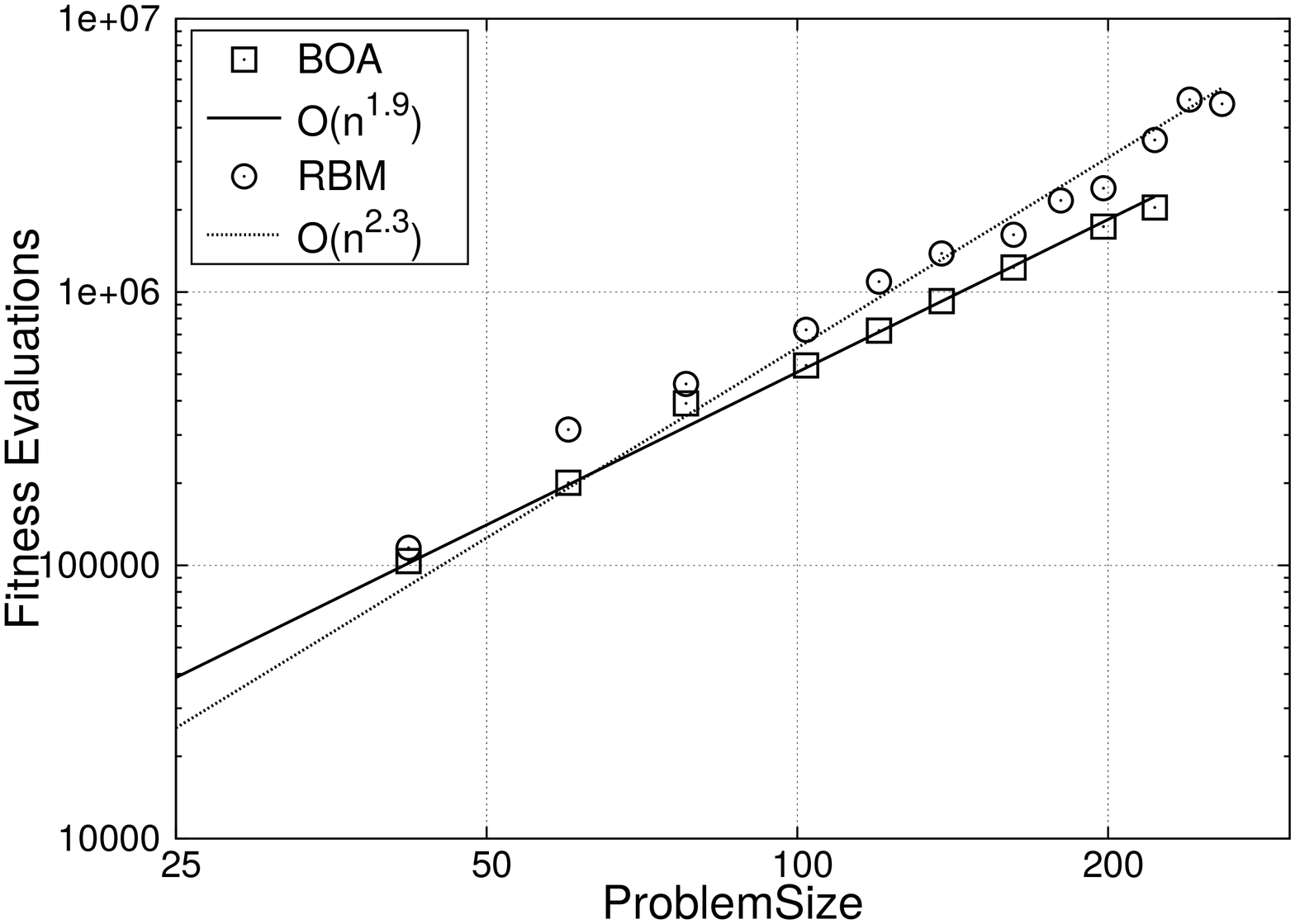}
      \includegraphics[width=0.45\linewidth]{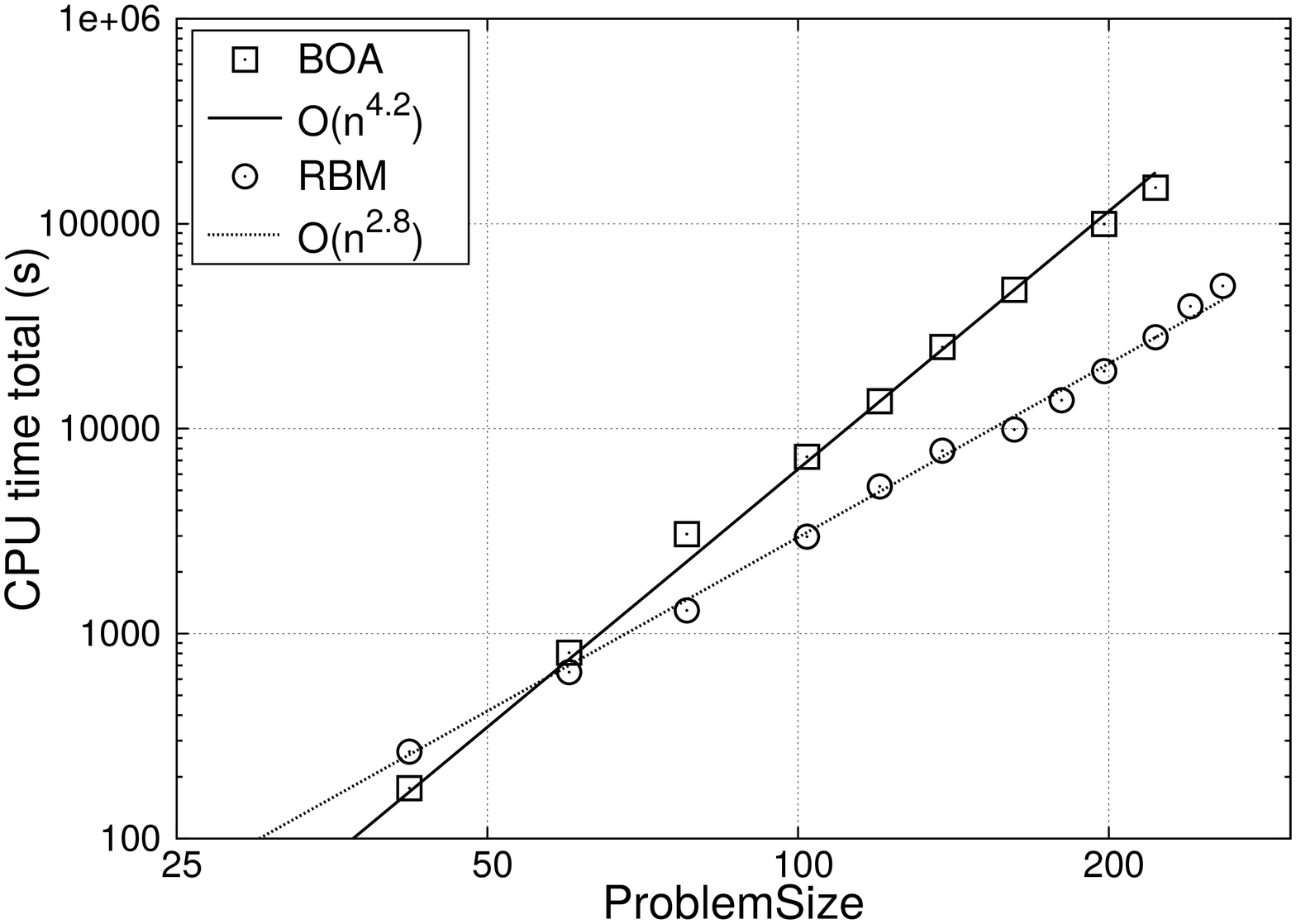}
    }
    \caption{Number of fitness evaluations (left) and CPU time (right) over problem size $l$ for deceptive traps with $(a)\;k=4, (b)\;k=5 \text{, and } (c)\;k=6$. \label{fig-traps}}
  \end{center}
\end{figure}

\begin{figure}
  \begin{center}
    \subfigure[$k=2$]
    {
      \includegraphics[width=0.45\linewidth]{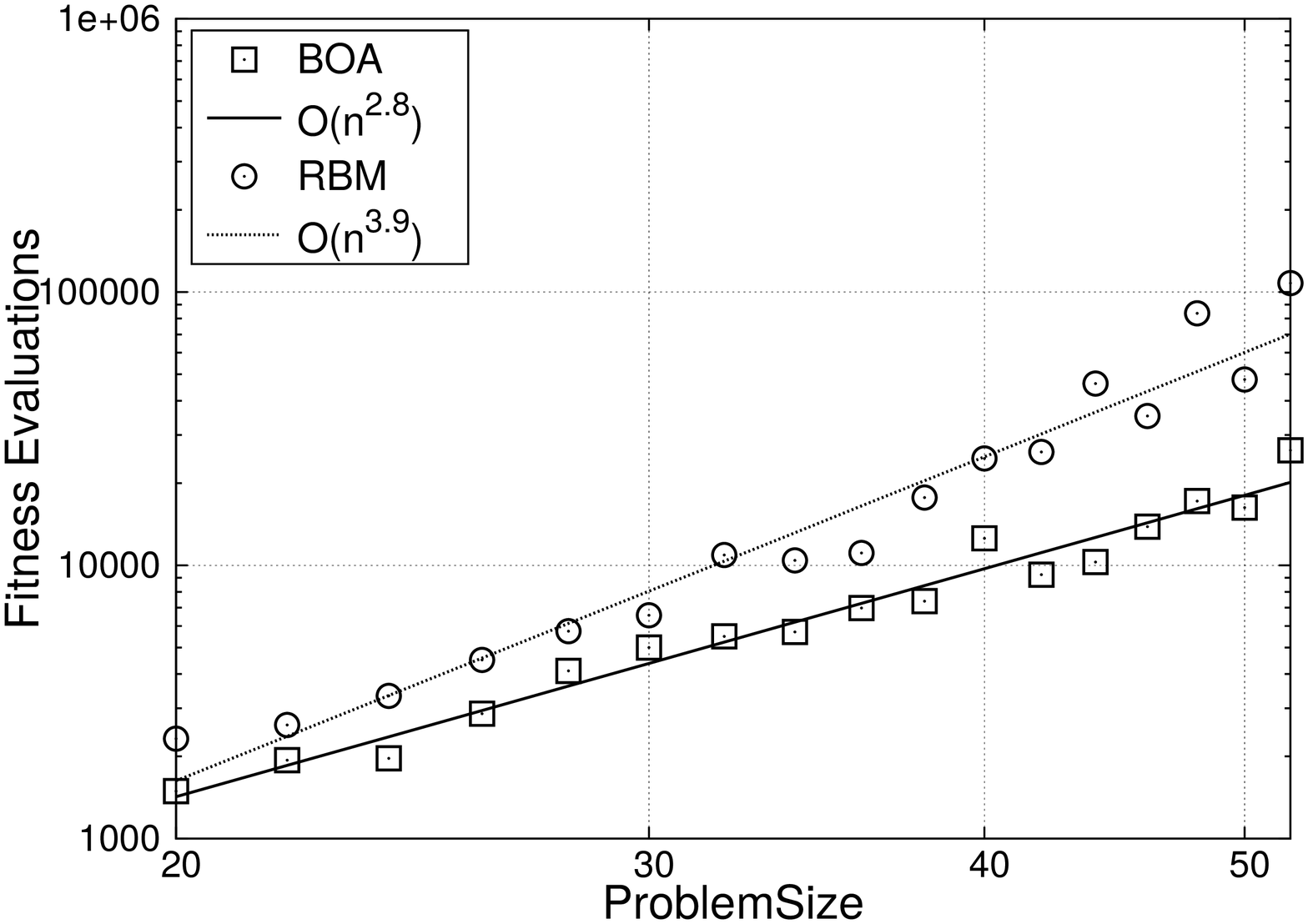}      \label{fig-nk-k2-eval}
      \includegraphics[width=0.45\linewidth]{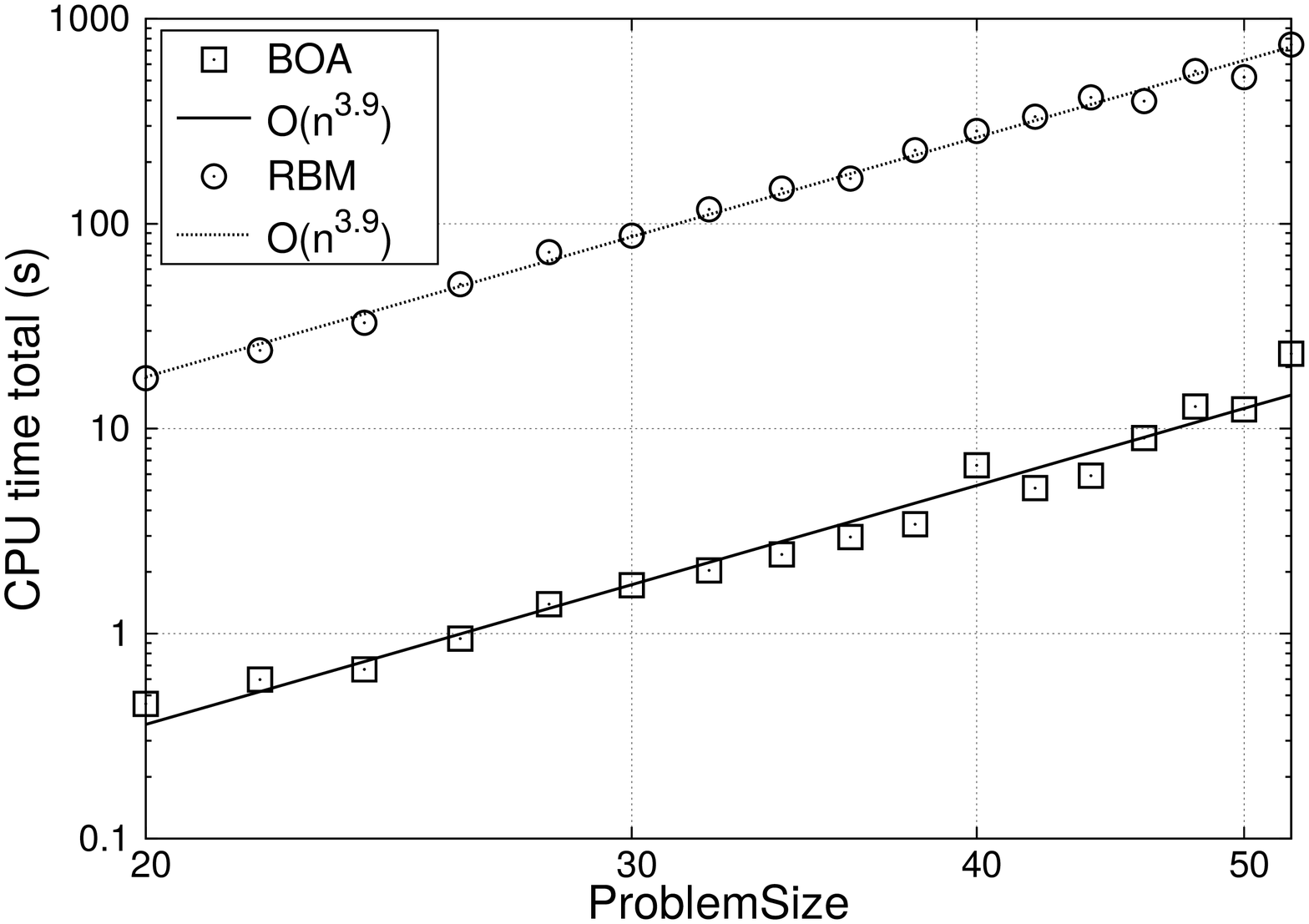}\label{fig-nk-k2-cpu}
    }
    \subfigure[$k=3$]
    {
      \includegraphics[width=0.45\linewidth]{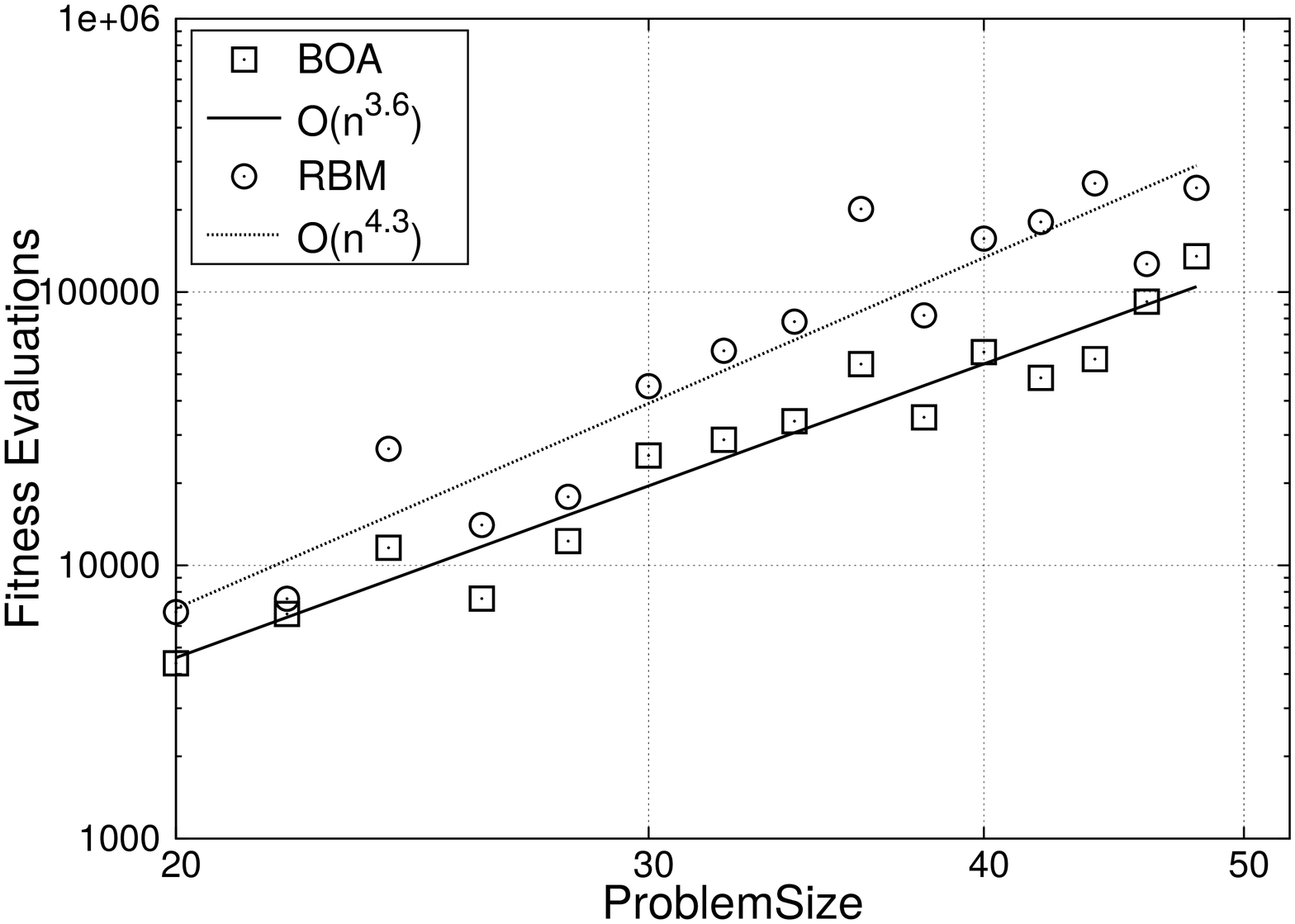}
      \label{fig-nk-k3-eval}
      \includegraphics[width=0.45\linewidth]{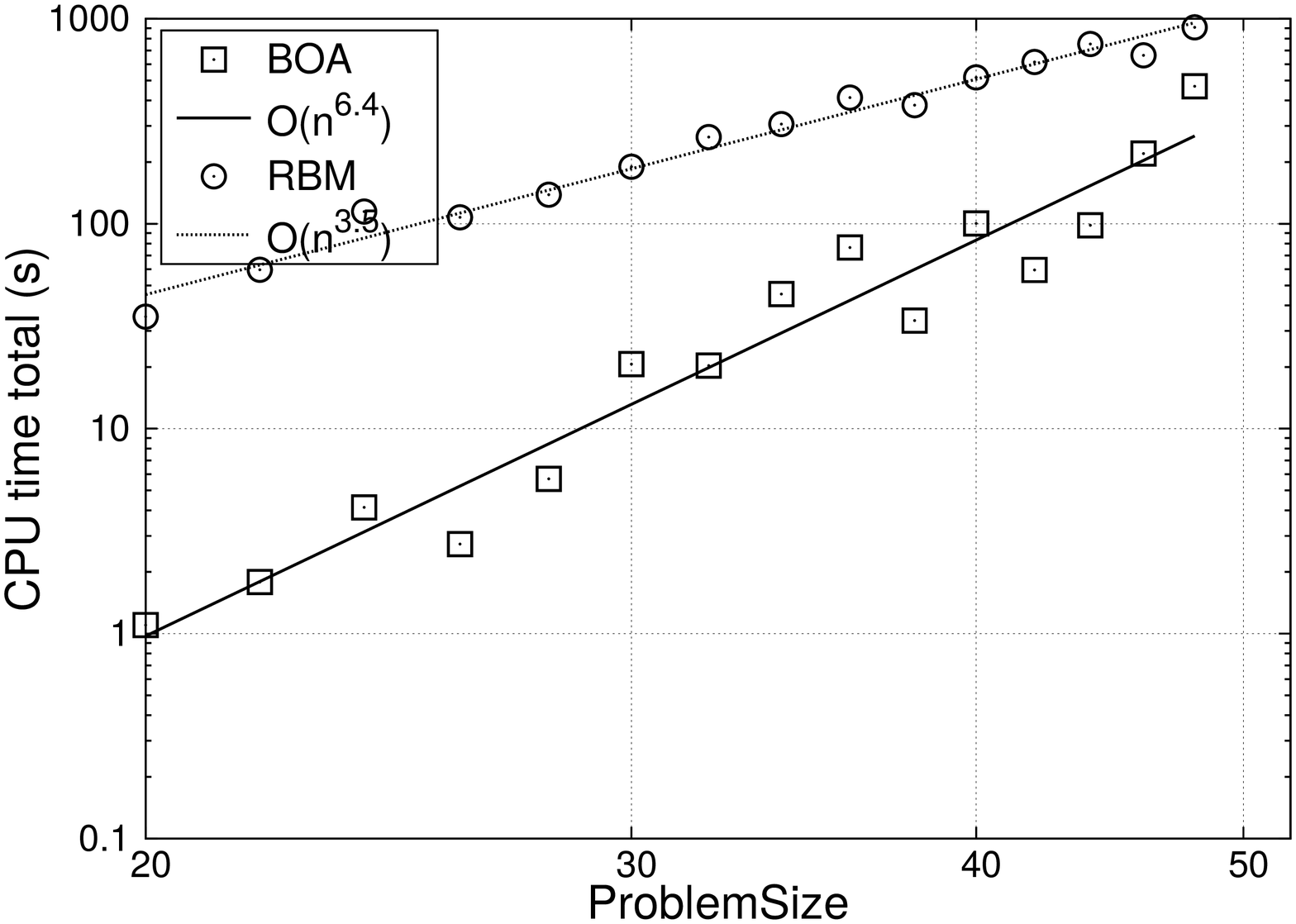}
      \label{fig-nk-k3-cpu}
    }
    \subfigure[$k=4$]
    {
      \includegraphics[width=0.45\linewidth]{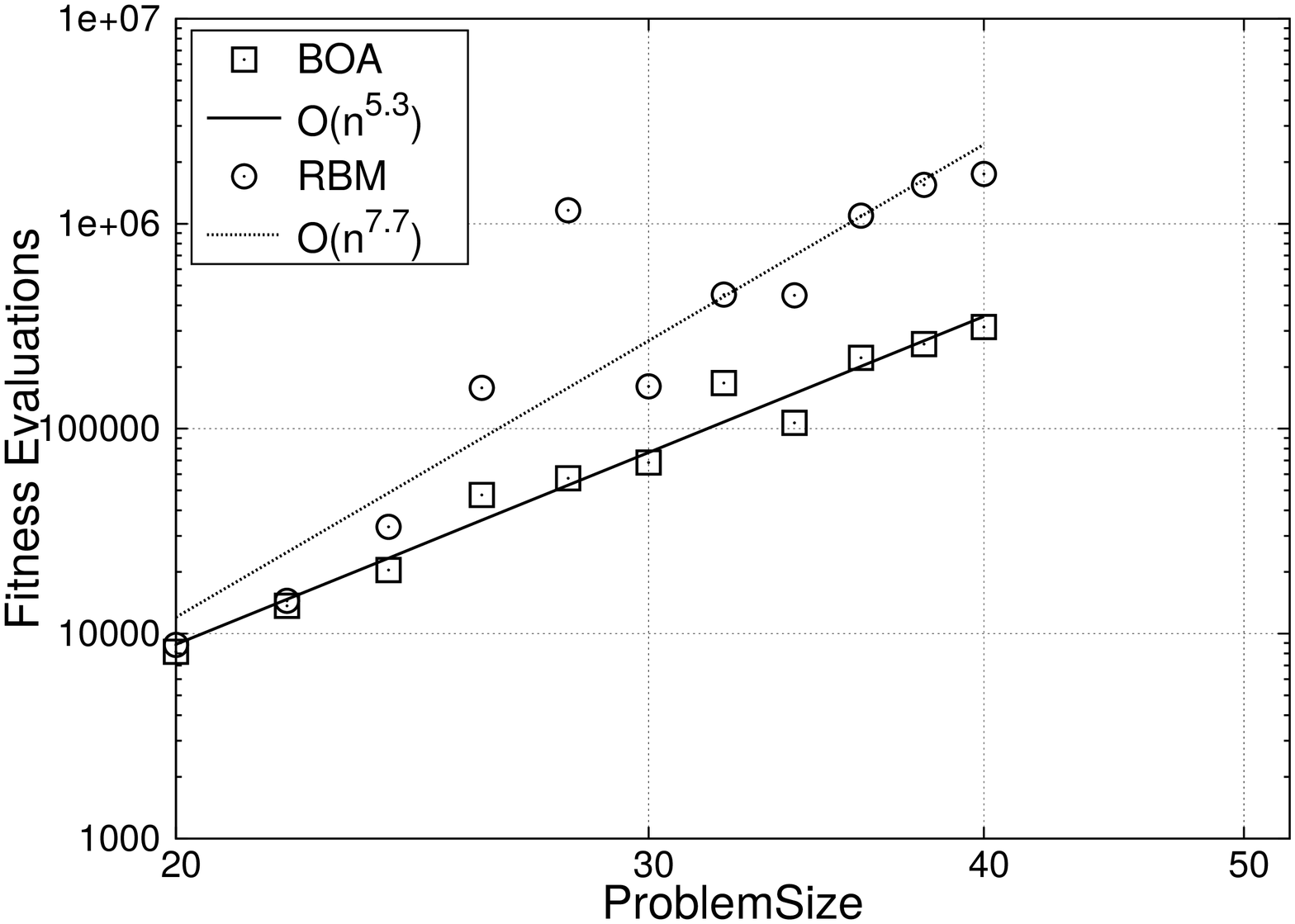}                \label{fig-nk-k4-eval}
      \includegraphics[width=0.45\linewidth]{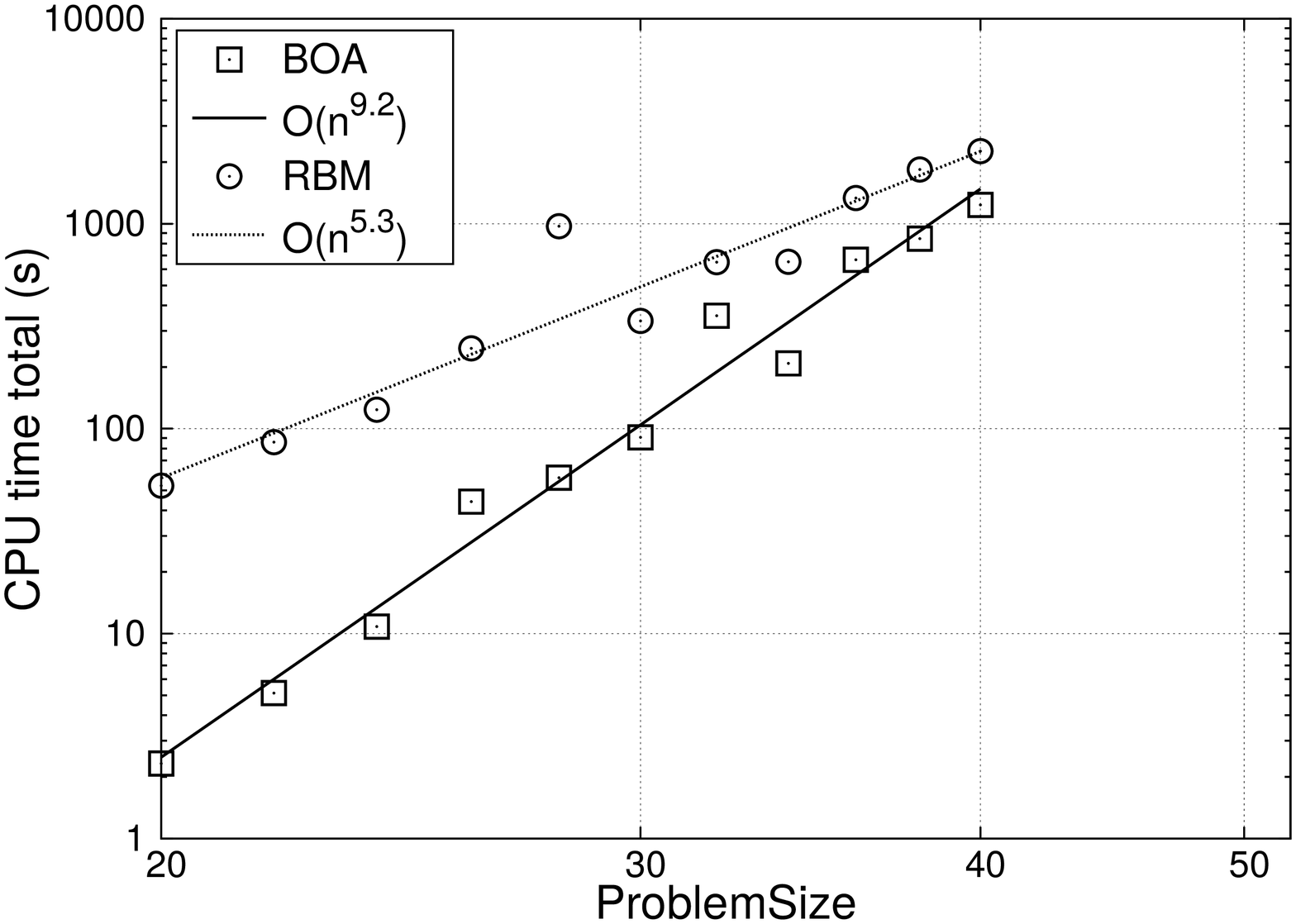}                \label{fig-nk-k4-cpu}
    }
    \subfigure[$k=5$]
    {
      \includegraphics[width=0.45\textwidth]{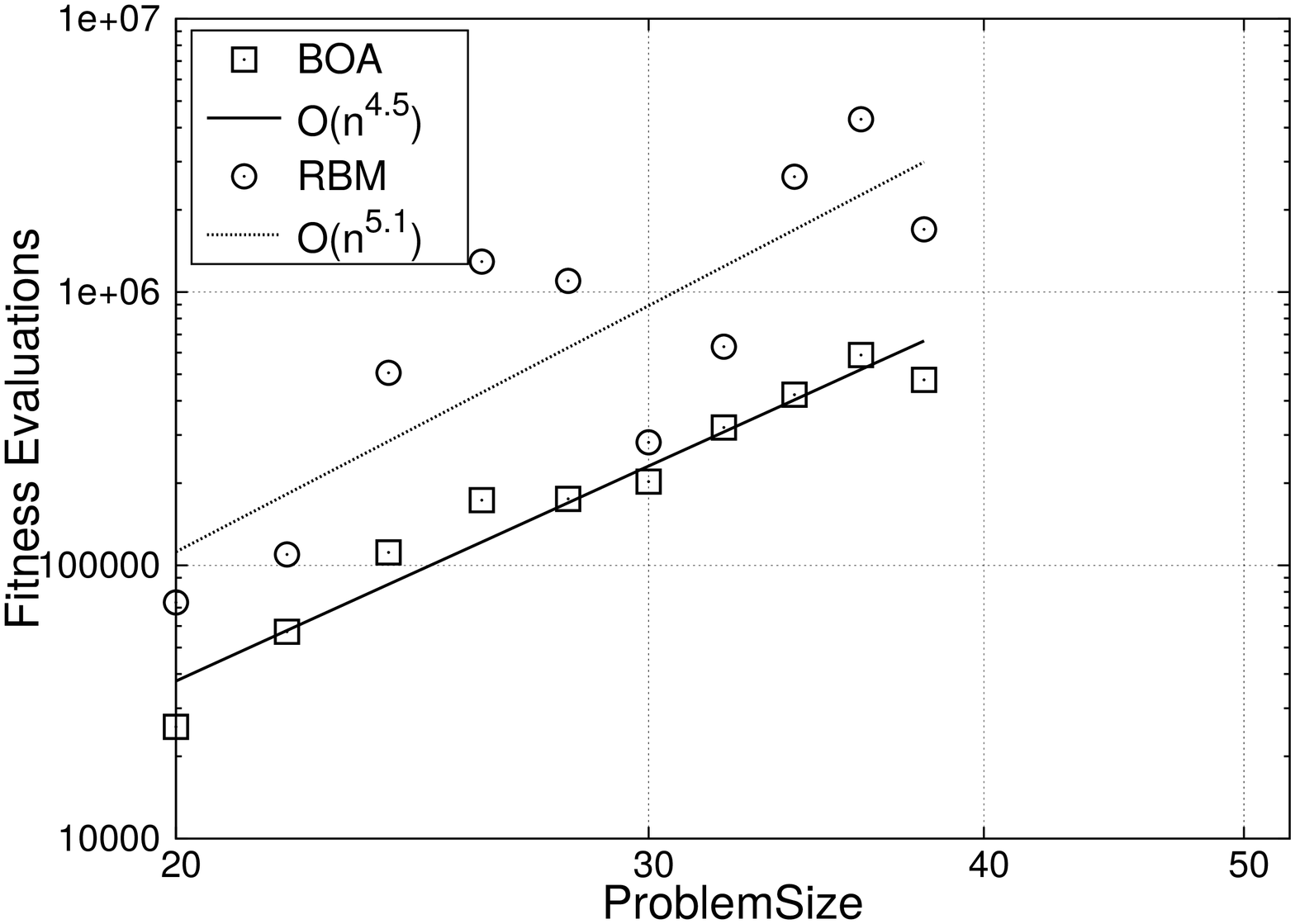}                \label{fig-nk-k5-eval}
      \includegraphics[width=0.45\textwidth]{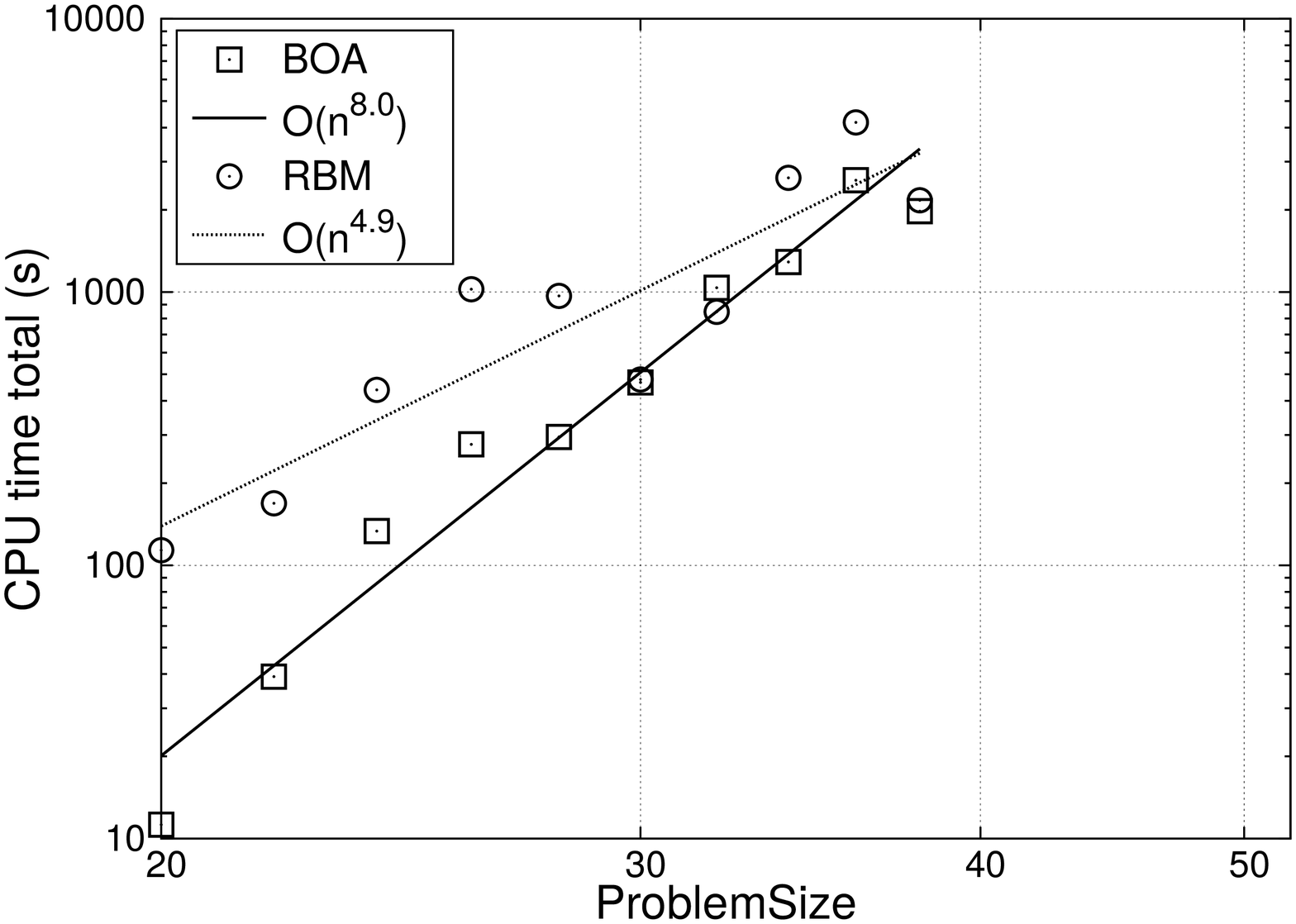}    }
    \caption{Number of evaluations (left) and CPU time (right) over problem size $l=N$ for NK landscapes with $(a)\;k=2, (b)\;k=3, (c)\;k=4 \text{, and } (d)\;k=5$. \label{fig-nk}}
  \end{center}
\end{figure}

We report the performance of RBM-EDA and BOA on onemax (Figure~\ref{fig-onemax}), concatenated traps (Figure~\ref{fig-traps}), and NK landscapes (Figure~\ref{fig-nk}). The figures have a log-log scale. Straight lines indicate polynomial scalability.
Each figure shows the average number of fitness evaluations (left-hand side) and the overall CPU time (right-hand side) required until the optimal solution was found. The figures also show regression lines obtained from fitting a polynomial to the raw results (details are in Table \ref{table-fitnesseval}).

First, we study the number of fitness evaluations required until the optimal solution was found (Figures \ref{fig-onemax}-\ref{fig-nk}, left).
For the onemax problem, RBM-EDA needed fewer fitness evaluations than BOA (Figure~\ref{fig-onemax}, left-hand side) and had a slightly lower complexity ($O(n^{1.3})$ vs.~$O(n^{1.6})$).
For concatenated traps, BOA needed less fitness evaluations (Figure~\ref{fig-traps}, left-hand side). As the problem difficulty increased (larger $k$), the  performance of the two approaches became more similar.
The complexity of BOA was slightly lower than the one of RBM-EDA (around $O(n^{1.8})$ versus $O(n^{2.3})$ for larger problems). 
The situation was similar for NK landscapes, where BOA needed fewer fitness evaluations and scaled better than RBM-EDA (Figure~\ref{fig-nk}, left-hand side).

We now discuss average total CPU times (Figures \ref{fig-onemax}-\ref{fig-nk}, right-hand side). Besides the time required for fitness evaluation, this includes time spent for model building, sampling, and selection.
For the onemax problem, RBM-EDA was faster than BOA and had a lower time complexity (Figure~\ref{fig-onemax}, right-hand side). 
For deceptive trap functions, BOA was faster on small problem instances, but its time complexity was larger that that of RBM-EDA (Figure~\ref{fig-traps}, right-hand side). Hence, the absolute difference in computation time became smaller for larger and more difficult problems.
For traps of size $k=5$, RBM-EDA was faster for problems with more than 180 decision variables (36 concatenated traps). For traps of size $k=6$, RBM-EDA was already faster for problems with more than 60 decision variables (10 concatenated traps).
BOA's  time complexity increased slightly with higher $k$ (from $O(n^{3.9})$ for $k=4$ to $O(n^{4.2})$ for $k=6$). In contrast, the time complexity of RBM-EDA remained about the same (between $O(n^{3.0})$ and $O(n^{2.8})$). 

The results for NK landscapes were qualitatively similar (Figure~\ref{fig-nk}, right-hand side). BOA was faster than EDA-RBM, but its computational effort increased stronger with $n$. Therefore the computation times became similar for difficult and large problems (cf. results for $k=5,n \in {30,32,38}$).

\begin{figure}
  \begin{center}
    \subfigure[Onemax\label{fig-time-onemax}]
    {
      \includegraphics[width=0.45\linewidth]{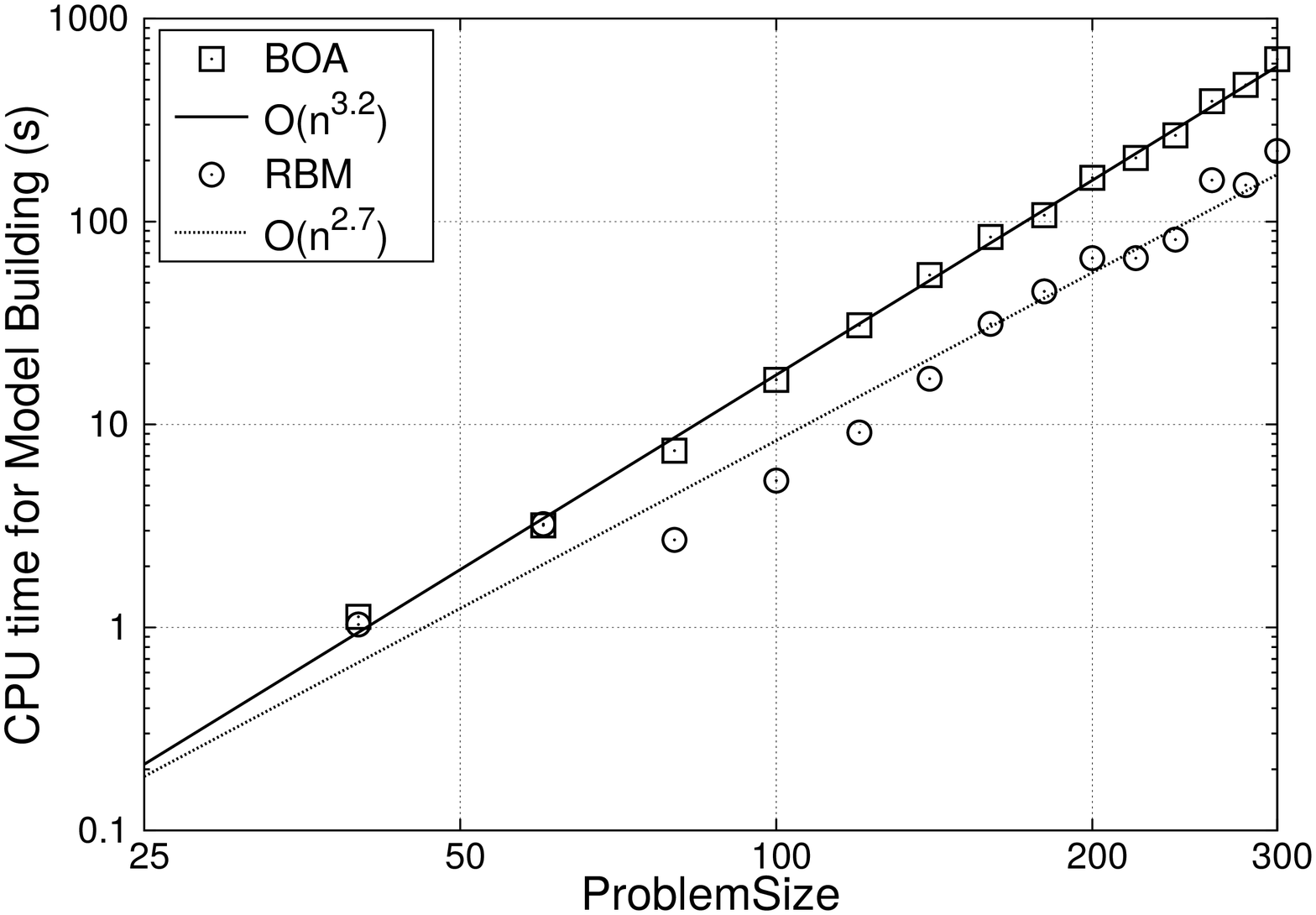}\label{fig-time-onemax-abs}
      \includegraphics[width=0.46\linewidth]{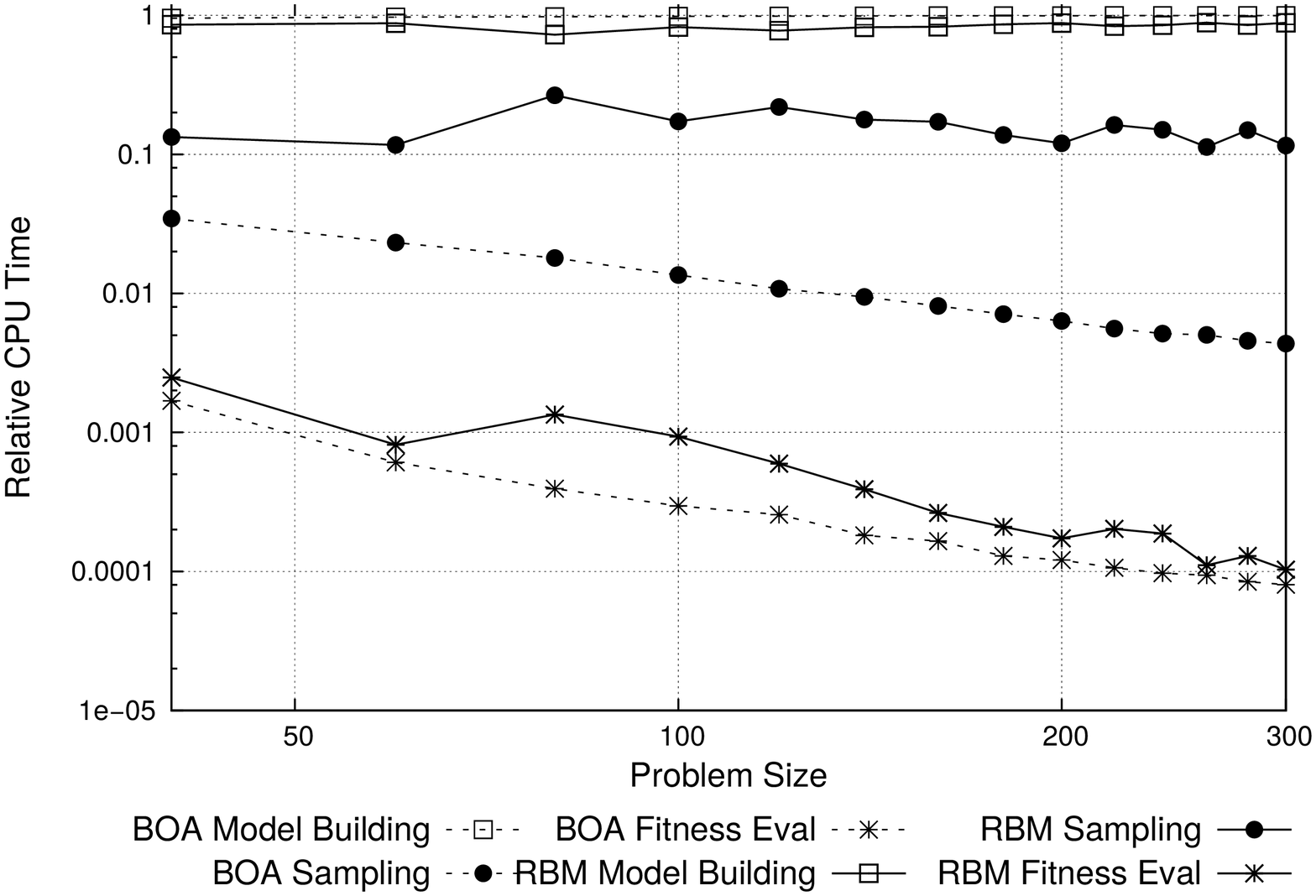}\label{fig-time-onemax-rel}
    }
    \subfigure[Concatenated traps, $k=5$\label{fig-time-5-traps}]
    {
      \includegraphics[width=0.45\linewidth]{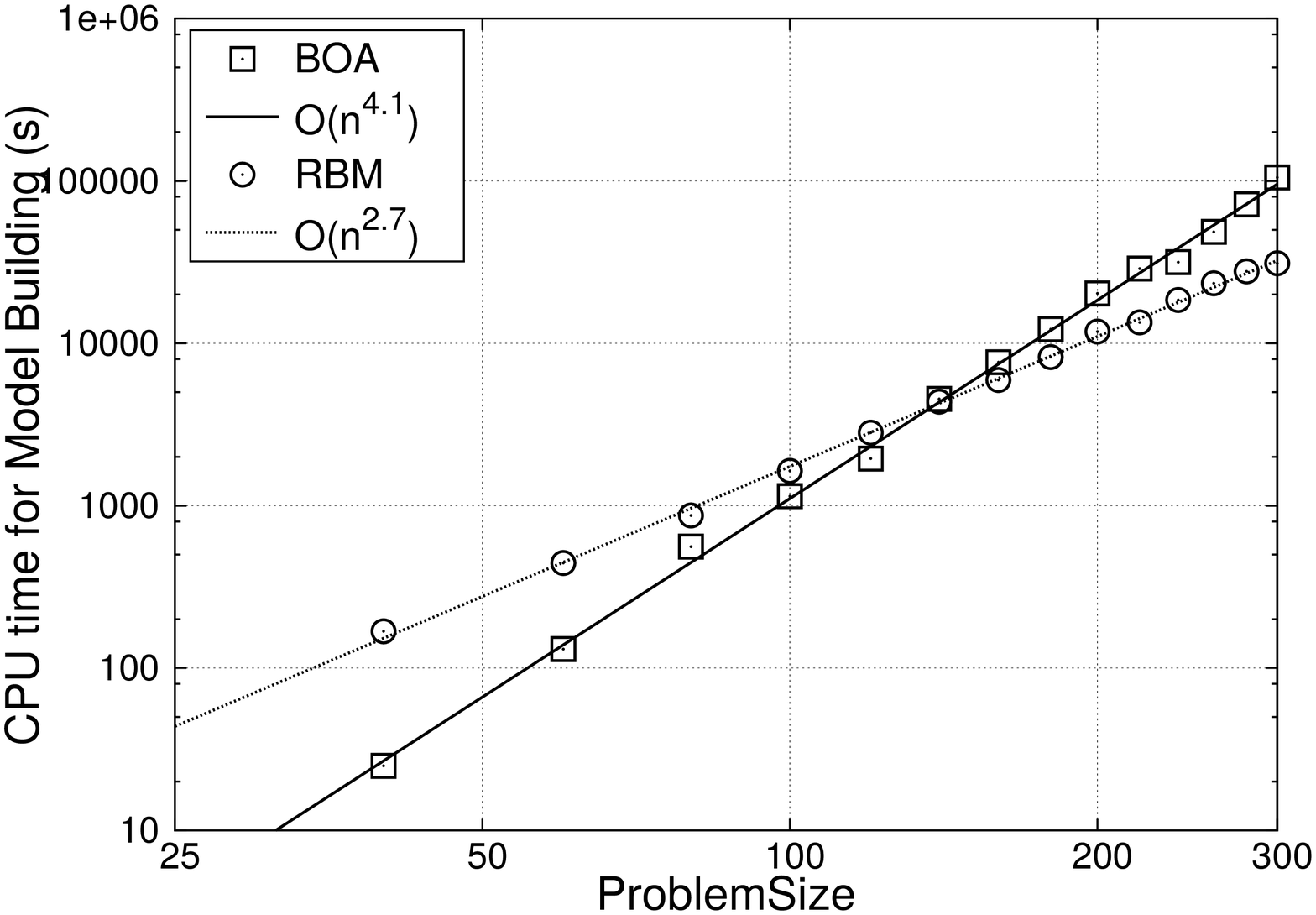}\label{fig-time-5-traps-abs}
      \includegraphics[width=0.46\linewidth]{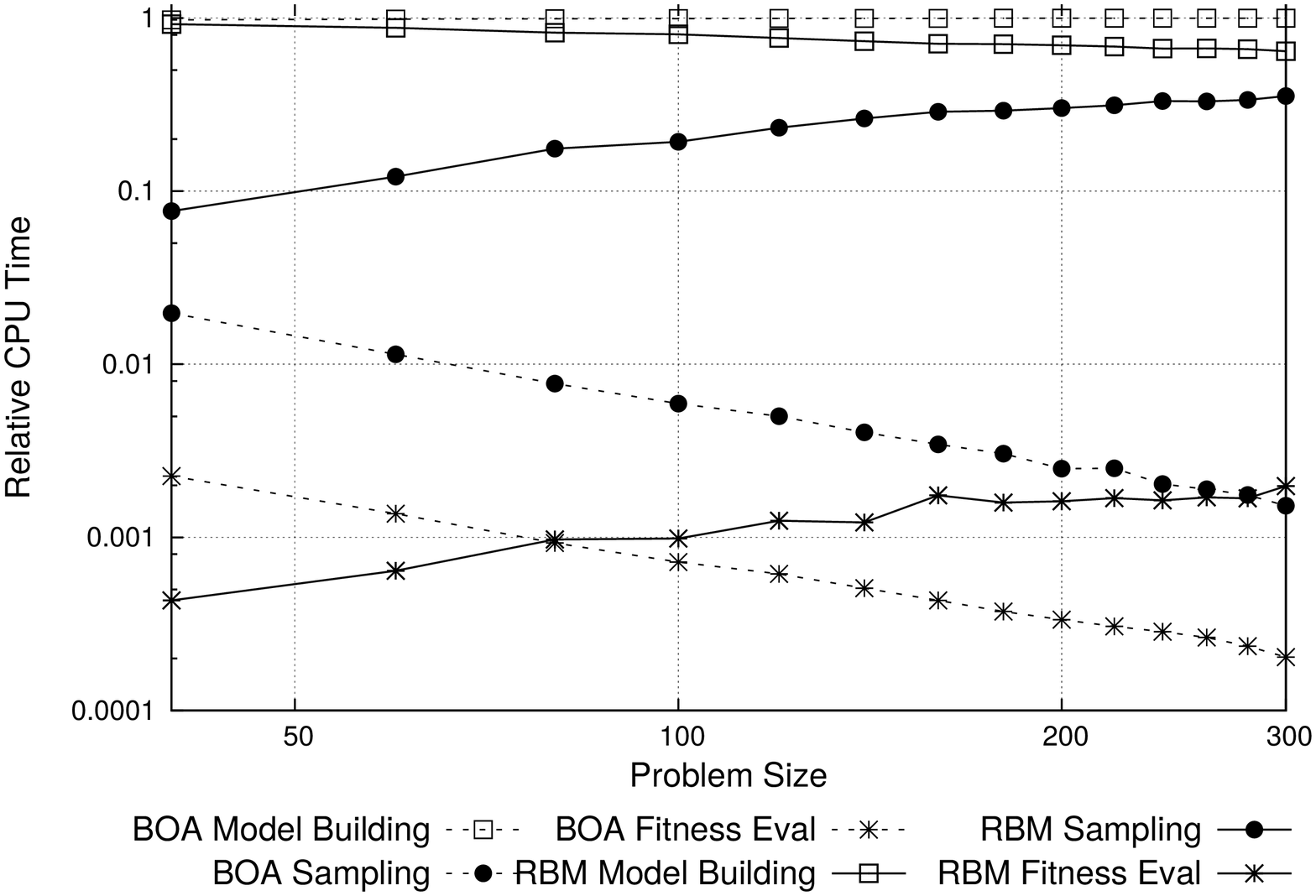}\label{fig-time-5-traps-rel}
    }
    \subfigure[NK landscapes, $k=5$\label{fig-time-nk-k5}]
    {
	  \includegraphics[width=0.45\linewidth]{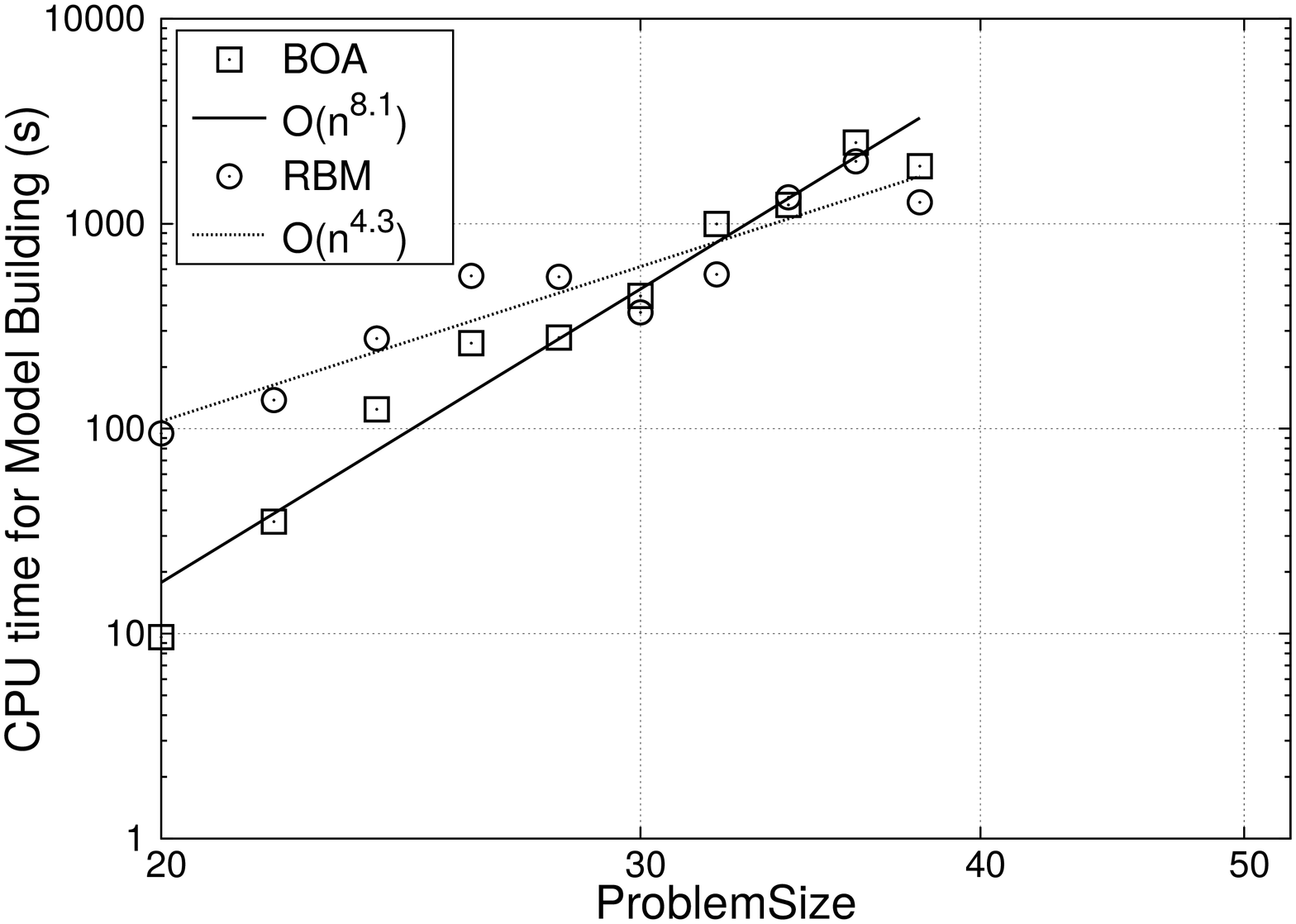}\label{fig-time-nk-k5-abs}
      \includegraphics[width=0.46\linewidth]{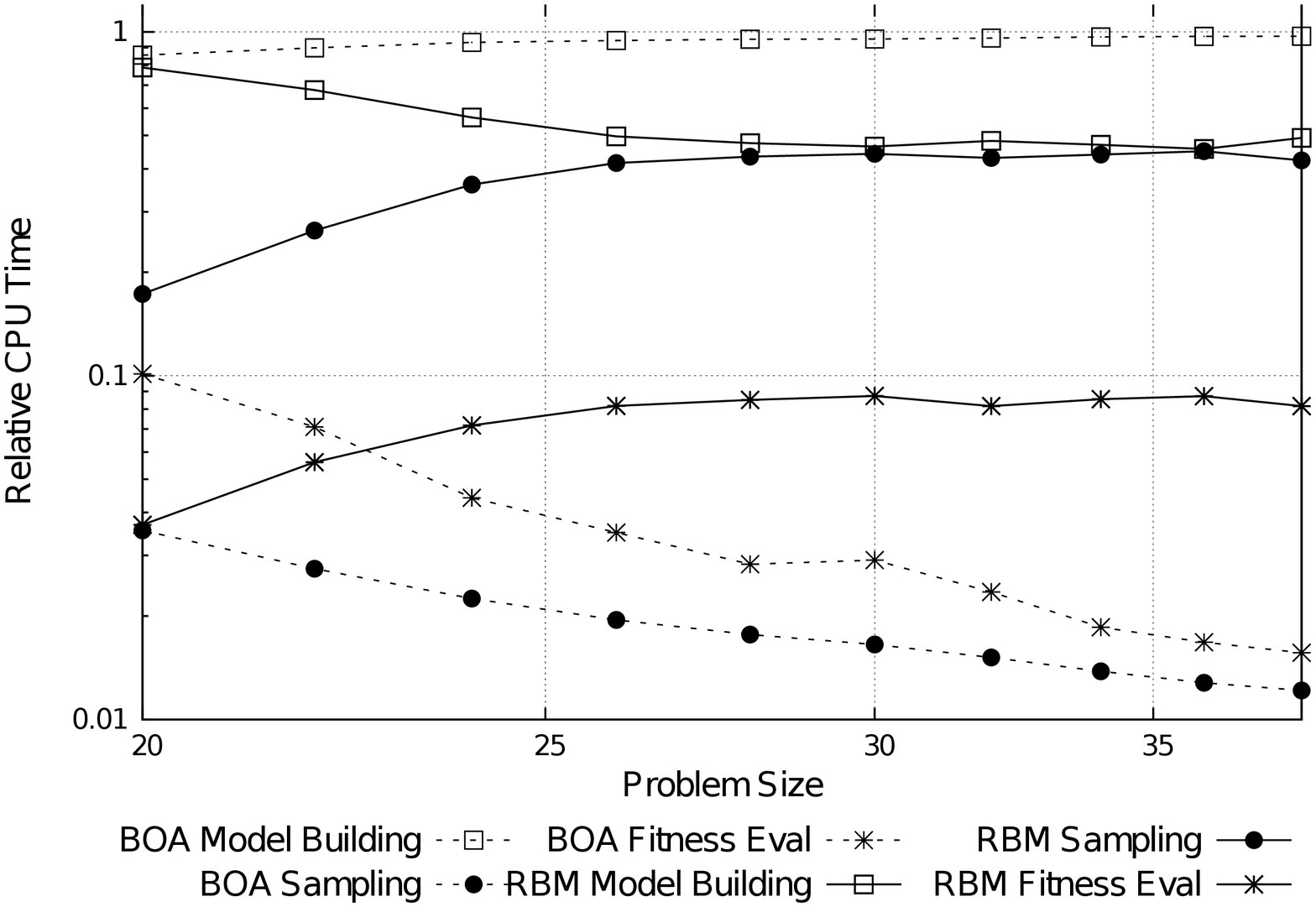}\label{fig-time-nk-k5-rel}
    }
    \caption{Relative CPU times for model building, sampling, and fitness evaluation. \label{fig-times}}
  \end{center}
\end{figure}
The results indicate that although RBM-EDA needed a higher number of fitness evaluations, its overall computational effort was about similar or even lower than BOA, especially for difficult problems. For this reason, we analyze in greater detail the differences in computations effort (CPU time) between BOA and RBM-EDA. Figure \ref{fig-times} exemplarily shows the absolute CPU times for model building (left-hand side) and relative CPU times necessary for model building, sampling, and fitness evaluation (right-hand side) for onemax, concatenated 5-traps and NK landscapes with $k=5$ for various problem sizes\footnote{We omitted the relative CPU times for selection in order to increase the readability of Figure \ref{fig-times}. By definition, they are proportional to the CPU time for fitness evaluations, and, in absolute numbers, negligible.}$^{,}$\footnote{The results for concatenated traps of size $k\in(4,6)$ and NK landscapes of size $k\in(3,4,6)$ were qualitatively similar.}.

First, we study the absolute CPU times for model building (Figure \ref{fig-times}, left-hand side). For all three problem types, RBM-EDA's model building had a lower CPU time complexity (Onemax: $O(n^{2.7})$ vs. $O(n^{3.2})$, 5-traps: $O(n^{2.7})$ vs. $O(n^{4.1})$, NK landscapes $k=5$: $O(n^{4.3})$ vs. $O(n^{8.1})$). Model building was faster for RBM-EDA than for BOA for all onemax problems, for concatenated 5-traps of size $>=140$ and for NK-$k=5$ problems of size $\in(30,32,36,38)$. Note that this was despite the fact that the population size, i.e., the training set, was usually bigger for RBM-EDA.

Second, we analyze the relative amount of CPU time that each EDA part used (Figure \ref{fig-times}, right-hand side). For BOA, model building dominated the CPU usage with more than 95\% of the total CPU time for all problems. Furthermore, with growing problem size, the share monotonously increased  (95.2-99.5\% for onemax, 97.4-99.8\% for 5-traps, 85.6-97.2\% for NK-$k=5$). The relative times for sampling and fitness were very low. Furthermore, their shares of the total time decreased with growing problem size, despite the growing population sizes. 
In contrast, the RBM spent a significantly smaller fraction of the total time for model building. Also, with growing problem size, this fraction decreased or stayed relatively constant (85.7-88.3\% for onemax, 92.2-64.3\% for 5-traps, 78.8-49.2\% for NK-$k=5$). Correspondingly, the CPU time fractions for sampling and fitness evaluations increased with growing problem size. Hence, the total CPU time for RBM-EDA was much less dominated by model building\footnote{Note that in this comparison, each algorithm used the population sizes from its own bisection run, i.e., the population size for BOA was usually smaller. If we used the same population sizes, we expect the time dominance of model building in BOA to be even greater.}. 

In summary, we found that the performance of RBM-EDA was competitive with the state-of-the art, especially for large and difficult problem instances. This may be surprising as the number of fitness evaluations necessary to solve a problem was, in most problems, higher for RBM-EDA than for BOA.
Even more, the computational effort in terms of fitness evaluations grew faster with the problem size $n$. The higher number of fitness evaluations used by RBM-EDA indicates that the statistical model created during training in RBM-EDA was less accurate than the statistical model in BOA. However, from a computational perspective, building this accurate model in BOA was much more expensive than learning the more inaccurate model used in RBM-EDA. Furthermore, the time necessary for building  the model increased slower with increasing problem size than for BOA. Thus, the lower time effort necessary for learning a less accurate statistical model in RBM-EDA overcompensated for the higher number of fitness evaluations that were necessary to find optimal solutions. As a result, the CPU times for RBM-EDA were not only lower than BOA, they also increased more slowly with growing problem sizes. This was true especially for difficult and large problems.

\begin{table}
\footnotesize
\centering
    \caption{Approximated scaling behavior between the number of fitness evaluations and problem size, as well as CPU times and problem size}
    \begin{tabular}{l*4c}
    \toprule
        & \multicolumn{2}{c}{Fitness evaluations} & \multicolumn{2}{c}{CPU time}\\
        ~       & BOA                   & RBM               & BOA           & RBM   \\
        \midrule
        ONEMAX  & ${O(n^{1.3})}$ &$O(n^{1.6})$       & $O(n^{3.2})$  & ${O(n^{2.7}}$)\\
        4-TRAPS & ${O(n^{1.8})}$ & $O(n^{2.8})$      & $O(n^{3.9})$  & ${O(n^{3.0})}$\\
        5-TRAPS & ${O(n^{1.8})}$ & $O(n^{2.4})$      & $O(n^{4.0})$  & ${O(n^{2.8})}$\\
        6-TRAPS & ${O(n^{1.9})}$ & $O(n^{2.3})$ & $O(n^{4.2})$ & ${O(n^{2.8})}$\\
        NK k=2& ${O(n^{2.8})}$ & $O(n^{3.9})$ & ${O(n^{3.9}})$ & ${O(n^{3.9})}$\\
        NK k=3& ${O(n^{3.6})}$ & $O(n^{4.3})$ & $O(n^{6.4})$ & ${O(n^{3.5})}$\\
        NK k=4& ${O(n^{5.3})}$ & $O(n^{7.7})$ & $O(n^{9.2})$ & ${O(n^{5.3})}$\\
        NK k=5& ${O(n^{4.5})}$ & $O(n^{5.1})$ & $O(n^{8.0})$ & ${O(n^{4.9})}$\\
     \bottomrule
    \end{tabular}
    \label{table-fitnesseval}
\end{table}

\section{Summary and Conclusions}
\label{conclusion}
We carried out an in-depth experimental analysis of using a Restricted Boltzmann Machine within an Estimation of Distribution Algorithm for combinatorial optimization. We tested RBM-EDA on standard binary benchmark problems: onemax, concatenated deceptive traps and NK landscapes of different sizes. We carried out a scalability analysis for the number of fitness evaluations and the computation times required to solve the problems to optimality. We compared our results to those obtained from the Bayesian Optimization Algorithm,  a state-of-the-art method.

Our experimental results suggest that RBM-EDA is competitive with the state-of-the-art. We observed that it was less efficient in terms of fitness evaluations, both in absolute numbers and in complexity. However, the estimated complexity of probabilistic model building in RBM-EDA was lower than in BOA. RBM-EDA was able to build probabilistic models much faster than BOA if the problem is large. This caused smaller total runtimes for RBM-EDA than for BOA, especially if the problem instance was large and difficult (cf. results for concatenated traps with $l>60, k=5$, traps with $l>10,k=6$, or NK landscape with $N=5,k=32$). In sum, RBM-EDA can be an alternative if problems are large and difficult and the computational effort for fitness evaluations is low. This highlights the potential of using generative neural networks for combinatorial optimization.

Another advantage of using neural networks in EDAs is that RBMs can be parallelized without many of the problems that occur when parallelizing other EDAs. Parallelizing an RBM-EDA on a Graphics Processing Unit leads to massive speed-ups by a factor of 200 and more, compared to optimized CPU code \citep{probst2014a}. 

There are multiple directions for further research. We demonstrated that RBMs are useful in EDAs, but fine-tuning the parameters could improve the RBM's model quality, possibly leading to further performance improvements. It might also be beneficial to stack RBMs on multiple layers and to use such deep systems for solving hierarchical problems.

\bibliographystyle{elsarticle-harv}

\end{document}